**Highlights**

- A unified parameter-free framework eliminates fuzzification dependency through theoretically grounded entropy regularization with convergence guarantees.

- Hierarchical dimensionality reduction operates simultaneously at feature and view levels, achieving up to 97% computational efficiency gains while preserving clustering integrity.

- Signal-to-noise ratio based regularization ($\delta_j^h = \frac{\bar{x}_j^h}{(\sigma_j^h)^2}$) provides principled feature weighting without manual intervention.

- Comprehensive evaluation across five diverse domains demonstrates consistent superiority over 15 state-of-the-art methods from centralized and federated paradigms.

- Automatic identification of critical view combinations enables feature space reduction to as low as 0.45% of original dimensionality while maintaining clustering quality.

Graphical Abstract

**Parameter-free entropy-regularized multi-view clustering with hierarchical feature selection**

Kristina P. Sinaga, Sara Colantonio, Miin-Shen Yang

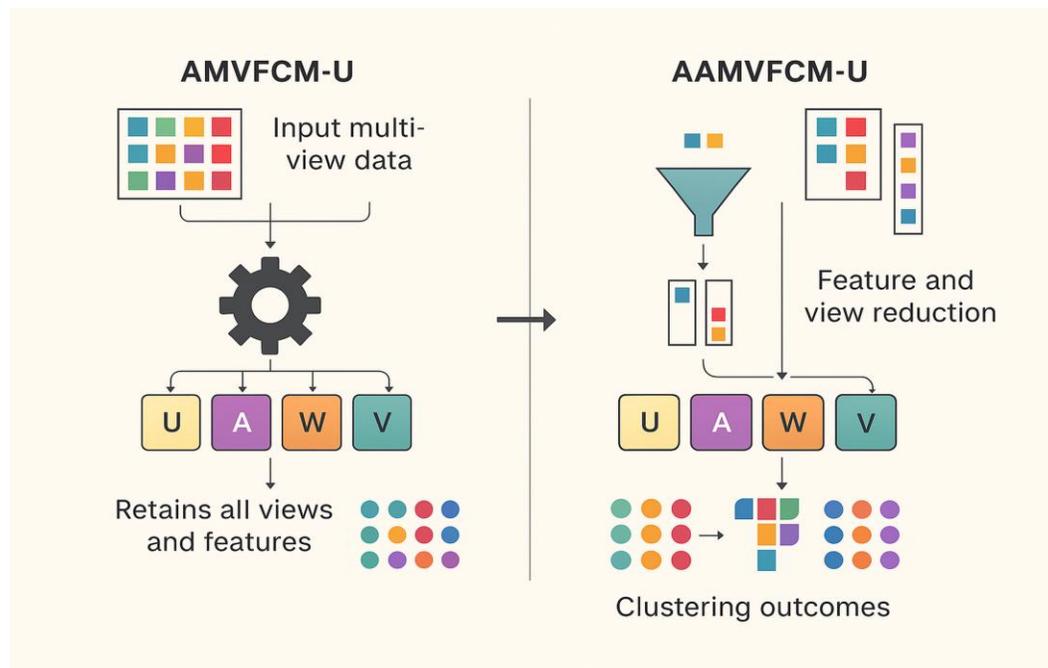

# Parameter-free entropy-regularized multi-view clustering with hierarchical feature selection

Kristina P. Sinaga[a,*], Sara Colantonio[a], Miin-Shen Yang[b]

[a]*Institute of Information Science and Technologies of the National Research Council of Italy (ISTI-CNR), Via G. Moruzzi, 1, 56124 Pisa, Italy*
[b]*Department of Applied Mathematics, Chung Yuan Christian University, Taoyuan 32023, Taiwan*

**Abstract.**

Multi-view clustering faces critical challenges in automatically discovering patterns across heterogeneous data while managing high-dimensional features and eliminating irrelevant information. Traditional approaches suffer from manual parameter tuning and lack principled cross-view integration mechanisms. This work introduces two complementary algorithms: AMVFCM-U and AAMVFCM-U, providing a unified parameter-free framework. Our approach replaces fuzzification parameters with entropy regularization terms that enforce adaptive cross-view consensus. The core innovation employs signal-to-noise ratio based regularization $\left( \delta_j^h = \dfrac{\overline{x}_j^h}{(\sigma_j^h)^2} \right)$ for principled feature weighting with convergence guarantees, coupled with dual-level entropy terms that automatically balance view and feature contributions. AAMVFCM-U extends this with hierarchical dimensionality reduction operating at feature and view levels through adaptive thresholding $\left( \theta^{h^{(t)}} = \dfrac{d_h^{(t)}}{n} \right)$. Evaluation across five diverse benchmarks demonstrates superiority over 15 state-of-the-art methods. AAMVFCM-U achieves up to 97% computational efficiency gains, reduces dimensionality to 0.45% of original size, and automatically identifies critical view combinations for optimal pattern discovery.

## 1. Introduction

Understanding complex data is crucial in today's data-driven world, and recent advancements in machine learning are significantly enhancing our ability to analyze and interpret this information. The development of new architectures for managing complex data is advancing rapidly. Particularly noteworthy is the use of mathematical optimization in unsupervised clustering algorithms, which ensures the removal of irrelevant data and outliers through techniques such as factorization, tensorization, and matrix embedding, all while preserving privacy. This requires modernizing the core elements of traditional methods. The original soft clustering approach was designed to process data from a single source [1]. In its traditional form, it processed input data by weighing feature components equally, allowing data points to hold the probabilities of belonging to predefined clusters. To address its limitations, advances in soft clustering have been made by employing feature weighting to better measure the significance of features in revealing clustering structures in input data [2,3]. Numerous studies have demonstrated the application of soft clustering based on feature weighting in critical domains [4,5]. The core concept involves modifying the mathematical formulations of the standard soft clustering approaches [6–10].

However, traditional mathematical optimization principles are becoming outdated as data-driven decision making increasingly relies on diverse resources distributed across multiple locations. We are in the process of transitioning from centralized to decentralized machine learning algorithms. Therefore, it is crucial to improve traditional unsupervised clustering algorithms, using their potential to uncover hidden patterns within large, complex datasets from various sources.

In 2004, Bickel et al. proposed a pioneering extension to enhance traditional k-means and expectation maximization (EM) for document clustering in a multi-view (MV) environment [11]. In 2009, Cleuzio et al. introduced multi-view clustering (MVC) based on standard fuzzy c-means (FCM) [12]. Since then, there has been a marked increase in research focusing on the identification of patterns in MV data, demonstrating widely applicable in real-world scenarios [13–15].

Multi-view FCM (MVFCM) as one of the MV learning versions of soft clustering algorithm families had been significantly expanded in non-federated and federated environments [16–22]. Unfortunately, the variable $m$ as a key factor that can have a significant impact on the result of cluster formation in these MVFCM clustering algorithms remains unsolved. Thus, we propose a new MVC based on standard FCM with an automation stage to detect relevancy information provided by data views without the need to be controlled by a fuzzification parameter. Our contribution also includes an exploration of methods to distinguish irrelevant features from relevant features by iteratively detecting and removing irrelevant features from the MV data. We devise an objective function for adaptive multi-view FCM clustering without fuzzy rules (AMVFCM-U) and provide its derivation and proof. We eventually propose two algorithms that incorporate an alternative to AMVFCM-U (AAMVFCM-U) to address the problem of feature reduction on data views. Both proposed methods are designed for horizontal and vertical data frames, where subsets of data views are initially shared with the same data points that represent multiple distinct feature components.

The paper is organized as follows. Section 2 examines soft clustering algorithms that utilize feature weights for single and MV data. Section 3 outlines the proposed AMVFCM-U and AAMVFCM-U algorithms. Section 4 details the experiments and their results, while the conclusions are given in Section 5.

## 2. Literature Reviews

This section will focus on the basics and extensions of (weighted) soft clustering in single-view and multi-view environments with (out) utilizing tensor decomposition. The discussion starts with (weighted) FCM in single-view environments and ends with multi-view environments. For the reader to comprehend the concept, we provide summarizations of pioneering objective functions and their extensions Table 2-Table 7. To aid in readability, Table 1 provides a concise summary of the notation and symbols used in this study.

### 2.1. Weighted FCM

The FCM clustering algorithm was first introduced by Bezdek with a priori assumption on the number of clusters [1]. FCM divides a set of $d$ dimensional vectors $X$ into $c$ clusters where $X$ represents the $i$-th sample for the $j$-th feature with the probability constraint of cluster membership in the range of zero to one. Several algorithms in this direction have been widely employed to segment multivariate image [23–28], cluster electroencephalography (EEG) signals [29,30], monitoring the conditions of a selected land surface [31–33], etc. In general, these typical FCM clustering algorithms that utilize all the data features are equally important, highly sensitive to outliers, and initializations.

Table 1:  Mathematical notation and symbols used in this study

| Symbol | Description |
|--------|-------------|
| **Dataset Parameters** | |
| $n$ | Number of data points |
| $s$ | Number of views |

| $d_h$ | Number of dimensions in the $h$-th view |
|---|---|
| $x_{ij}^h$ | Value of the $j$-th feature of the $i$-th data point in the $h$-th view |
| **Clustering Parameters** | |
| $c$ | Number of clusters |
| $m$ | Fuzzification parameter (used in traditional FCM) |
| $t$ | Current iteration number |
| $\mu_{ik}$ | Membership degree of the $i$-th data point to the $k$-th cluster |
| $a_{kj}^h$ | Center of the $k$-th cluster for the $j$-th feature in the $h$-th view |
| **Weighting Parameters** | |
| $v_h$ | Weight of the $h$-th view |
| $w_j^h$ | Weight of the $j$-th feature in the $h$-th view |
| $\delta_j^h$ | Regularization parameter for the $j$-th feature in the $h$-th view |
| $\beta$ | Balancing parameter controlling the distribution of view weights $v_h$ |
| $\eta$ | Balancing parameter controlling the distribution of feature weights $w_j^h$ |

In 1999, Chen et al. [34] proposed a density-weighted FCM (DWFCM). DWFCM is the first approach in standard FCM for handling noisy data by quantifying density estimation of input data without considering measuring the contribution of feature components during clustering processes. In 2004, Wang et al. [2] introduced an automation to learn the importance of feature components during clustering processes. They proposed a feature-weight learning technique based on the gradient descent technique, called weighted FCM (WFCM). The WFCM automatically calculates the weights of all features and uses the weights to determine the importance of the features within a view. The WFCM can enhance the performance of the standard FCM and is robust to outliers. However, its performance depends on the selection of the distance metric and the fuzziness parameter.

Jing et al. [3] introduce entropy of features to present the importance of samples in a cluster based on standard k-means, called entropy weighting k-means (EWKM). The EWKM works well for analyzing sparse representations of clustered data in a high-dimensional space. However, its performance is controlled by a balancing parameter. A year later, Wang et al. [35] improved WFCM by introducing an approach to normalize input data first in the range of 0 to 1 to eliminate the effect of features with a wider variation range, called adaptive weights FCM (AWFCM). The concept of AWFCM is basically adopted from EWKM by reducing the effect of feature entropy on the distance formula of the standard FCM. To optimize clustering performance, AWFCM requires the best combination of the fuzzification parameter and the exponent parameter for feature weight. In the same year, Hadjahmadi et al. [36] introduced robust weighted FCM (RWFCM). The RWFCM adopted the idea of DWFCM to obtain robustness in the presence of outliers in clustered data.

Table 2: Categorized summary of (weighted feature-view) SVC algorithms in the literature

| Algorithm | Environment | Key Idea (s) | Parameter (s) |
|-----------|-------------|--------------|---------------|
| **Soft Clustering Methods** | | | |
| FCM [1] | Single-view; no weights | Classic fuzzy clustering that assigns probabilistic memberships to data points allowing them to belong to multiple clusters simultaneously. | Fuzzification $m$; clusters $c$ |
| DWFCM [34] | Single-view; no weights | Enhances FCM with noise/outlier robustness by incorporating local density information to adjust point influence. | Fuzzification $m$; clusters $c$; resolution $h$ |
| WFCM [2] | Single-view | Uses $\ell_2$-norm weighting to automatically determine importance of features during clustering. | Fuzzification $m$; clusters $c$ |
| AWFCM [35] | Single-view; adaptive weights | Introduces adaptive feature weights and supports normalized input to eliminate effects from features with wider variation. | Fuzzification $m$; clusters $c$; normalized input; exponent $\beta$ |
| IFWFCM_KD [6] | Single-view; kernel-based weights | Kernel-enhanced clustering that addresses non-linearity while improving resilience to noise and outliers. | Fuzzification $m$; clusters $c$ |
| **Hard Clustering Methods** | | | |
| EWKM [3] | Single-view; entropy-based weights | Employs entropy to determine feature relevance; effective for analyzing sparse representations in high-dimensional space. | Parameter $\alpha$; clusters $c$ |

Previous techniques were disabled to effectively handle non-linearity in clustered data. To address this limitation, Hathway et al. [7] proposed a weighted version of the non-Euclidean rational FCM algorithm (WNERFCM). Typically, they proposed three objective functions, namely relational FCM (RFCM), non-Euclidean RFCM (NERFCM), and weighted NERFCM (WNERFCM). The WNERFCM can be viewed as a feature reduction strategy without automation. In other words, users must manually reduce the features by referring to the midpoints of the histogram bins and the weights to the number of original objects in the bins. Unfortunately,

WNERFCM is costly when the number of dimensions of the input data is increasing. To overcome this issue, Xing et al. [6] developed an improved feature-weighted FCM with kernelized distance (IFWFCM_KD). IFWFCM_KD tackles the issue of nonlinearity in clustered data while simultaneously improving the standard FCM's resilience to noise and outliers. However, IFWFCM_KD is costly as it requires transforming the original input data into kernel space. After 2014, embedded feature weights became popular, massively developed, and implemented to solve various real-world issues [5,8,10,37–40].

Table 3:  Categorized objective functions for k-means and FCM-family algorithms in single-view environments

| Algorithm | Objective Function | Constraint (s) |
|---|---|---|
| **Standard FCM-based** | | |
| FCM [1] | $$J = \sum_{i=1}^{n} \sum_{k=1}^{c} \mu_{ik}^{m} \sum_{j=1}^{d} (x_{ij} - a_{kj})^2$$ | $\sum_{k=1}^{c} \mu_{ik} = 1, \mu_{ik} \in [0,1]$ |
| **Distance-Weighted Extensions** | | |
| DWFCM [34] | $$J = \sum_{i=1}^{n} \sum_{k=1}^{c} \mu_{ik}^{m} \sum_{j=1}^{d} w_j (x_{ij} - a_{kj})^2, w_j = \sum_{i=1, i' \neq i}^{n} \exp\left(-\frac{h \parallel x_i - x_{i'} \parallel}{\text{std}}\right)$$ | Same as FCM |
| WFCM [2] | $$J = \sum_{i=1}^{n} \sum_{k=1}^{c} \mu_{ik}^{m} \sum_{j=1}^{d} w_j^2 (x_{kj} - x_{ij})^2$$ | $\sum_{k=1}^{c} \mu_{ik} = 1, \mu_{ik} \in [0,1];$ $\sum_{j=1}^{d} w_j = 1, w_j \in [0,1]$ |
| WNERFCM [7] | $$J = \sum_{i=1}^{n} \sum_{k=1}^{c} \mu_{ik}^{m} \left\| w(x_i - a_k) \right\|_A^2$$ | Same as WFCM |
| **Entropy-Based Extensions** | | |
| EWKM [3] | $$J = \sum_{k=1}^{c} \left( \sum_{i=1}^{n} \mu_{ik} \sum_{j=1}^{d} w_{kj} (x_{ij} - a_{kj})^2 + \alpha \sum_{j=1}^{d} w_{kj} \log w_{kj} \right)$$ | $\sum_{k=1}^{c} \mu_{ik} = 1, \mu_{ik} \in \{0,1\};$ $\sum_{j=1}^{d} w_{kj} = 1, w_{kj} \in [0,1]$ |
| **Robust Weighting** | | |
| RWFCM [36] | $$J = \sum_{i=1}^{n} \sum_{k=1}^{c} \mu_{ik}^{m} \sum_{j=1}^{d} w_{kj} (x_{ij} - a_{kj})^2, w_{kj} = \left(\frac{1}{\|x_i - a_k\| + \eta}\right)^{\frac{2}{\alpha-1}}$$ | Same as FCM |

| AWFCM [35] | $$J = \sum_{i=1}^{n} \sum_{k=1}^{c} \mu_{ik}^{m} \sum_{j=1}^{d} w_{kj}^{\beta} (x_{ij} - a_{kj})^2$$ | $\sum_{k=1}^{c} \mu_{ik} = 1, \mu_{ik} \in [0,1];$ <br> $\sum_{j=1}^{d} w_{kj} = 1, w_{kj} \in [0,1]$ |
|---|---|---|
| **Kernel-Based Extensions** | | |
| IFWFCM_KD [6] | $$J = \sum_{i=1}^{n} \sum_{k=1}^{c} \mu_{ik}^{m} \left\| \Phi(\mathrm{diag}(w)x_i) - \Phi(\mathrm{diag}(w)a_k) \right\|$$ | Same as WFCM |

## 2.2. Weighted Multi-view FCM

Our research does not evaluate the significance of each view's feature component using feature entropy. Instead, we seek a global solution by integrating feature weights from different views with a supplementary membership entropy to mitigate the fuzzification guidelines of conventional FCM in centralized environments. Therefore, it is essential to explore different types of unsupervised multi-view clustering (MVC) algorithms and their evolution from centralized to decentralized approaches.

Table 4: Categorized objective functions for MV clustering (MVC) algorithms with or without view weights

| Algorithms | Objective Functions | Constraint (s) |
|---|---|---|
| **Category: Hard Clustering (Non-Fuzzy)** | | |
| MVKM-ED [41] | $$J = \sum_{h=1}^{s} (v_h)^{\alpha} \sum_{i=1}^{n} \sum_{k=1}^{c} \mu_{ik} \left\{ 1 - \exp\left( -\beta^h \sum_{j=1}^{d_h} \left( x_{ij}^h - a_{kj}^h \right)^2 \right) \right\}$$ | $\sum_{k=1}^{c} \mu_{ik} = 1, \mu_{ik} \in \{0,1\};$ <br> $\sum_{h=1}^{s} v_h = 1, v_h \in [0,1]$ |
| GKMVKM [41] | $$J = \sum_{h=1}^{s} (v_h)^{\alpha} \sum_{i=1}^{n} \sum_{k=1}^{c} \mu_{ik} \left\{ 1 - \left( \exp\left( -\beta^h \sum_{j=1}^{d_h} \left( x_{ij}^h - a_{kj}^h \right)^2 \right) \right)^p \right\}$$ | Same as MVKM-ED |
| **Category: Soft Clustering (Fuzzy)** | | |
| Co-FKM [12] | $$J = \sum_{h=1}^{s} \sum_{i=1}^{n} \sum_{k=1}^{c} (1-\alpha) \left( \mu_{ik}^h \right)^m + \frac{\alpha}{s-1} \left( \sum_{\substack{h'=1, \\ h' \neq h}}^{s} \left( \mu_{ik}^{h'} \right)^m \right) \sum_{j=1}^{d_h} \left( x_{ij}^h - a_{kj}^h \right)^2$$ | $\sum_{k=1}^{c} \mu_{ik}^h = 1, \mu_{ik}^h \in [0,1]$ |
| WV-Co-FCM [16] | $$J = \sum_{h=1}^{s} v_h \left( \sum_{i=1}^{n} \sum_{k=1}^{c} \left( \mu_{ik}^h \right)^m \left\| x_i^h - a_k^h \right\|^2 + \Delta_h \right) + \eta \sum_{h=1}^{s} v_h \log v_h$$ <br> $$\Delta_h = \sum_{i=1}^{n} \alpha_{ik}^h \sum_{k=1}^{c} \mu_{ik}^h \left( 1 - \left( \mu_{ik}^h \right)^{m-1} \right) - \sum_{i=1}^{n} \beta_{ik}^h \sum_{k=1}^{c} \mu_{ik}^h \left( 1 - \left( \mu_{ik}^h \right)^{m-1} \right)$$ | $\sum_{k=1}^{c} \mu_{ik}^h = 1, \mu_{ik}^h \in [0,1];$ <br> $\sum_{h=1}^{s} v_h = 1, v_h \in [0,1]$ |

| MinMax-FCM [18] | $J = \sum_{h=1}^{s}(v_h)^\alpha \sum_{i=1}^{n} \sum_{k=1}^{c} (\mu_{ik}^*)^m \left\| x_i^h - a_k^h \right\|^2$ | $\sum_{k=1}^{c} \mu_{ik}^* = 1, \mu_{ik}^* = [0,1];$ $\sum_{h=1}^{s} v_h = 1, v_h \in [0,1]$ |
|---|---|---|
| **Category: NMF-based Hard Clustering** | | |
| RMKMC [42] | $J = \sum_{h=1}^{s}(v_h)^\alpha \left\| (X^h)^T - U^*(A^h)^T \right\|_{2,1}$ | $\sum_{k=1}^{c} \mu_{ik}^* = 1, \mu_{ik}^* \in \{0,1\};$ $\sum_{h=1}^{s} v_h = 1, v_h \in [0,1]$ |
| MultiNMF [43] | $J = \sum_{h=1}^{s}\left\| (X^h)^T - U^h(A^h)^T \right\|_F^2 + \sum_{h=1}^{s} \eta_h \left\| U^h - U^* \right\|_F^2$ | $\sum_{k=1}^{c} \mu_{ik}^h = 1, \mu_{ik}^h \in [0,1]$ |
| **Category: Federated Learning (Decentralized)** | | |
| Fed-MVFCM [20] | $J = \sum_{l=1}^{L} \sum_{h=1}^{s}(v_{[l]}^h)^2 \sum_{i=1}^{n(l)} \sum_{k=1}^{c} (\mu_{[l]ik})^m \left\| x_{[l]i}^h - a_{[l]k}^h \right\|^2$ | $\sum_{k=1}^{c} \mu_{[l]ik} = 1, \mu_{[l]ik} \in [0,1];$ $\sum_{h=1}^{s} v_{[l]ji} = 1, v_{[l]ji} \in [0,1]$ |
| Fed-MVFPC [20] | $J = \sum_{l=1}^{L} \sum_{h=1}^{s}(v_{[l]}^h)^2 \sum_{i=1}^{n(l)} \sum_{k=1}^{c} (\mu_{[l]ik})^m \left\| x_{[l]i}^h - W^h p_k \right\|_2^2$ | $\sum_{k=1}^{c} \mu_{[l]ik} = 1, \mu_{[l]ik} \in [0,1];$ $\sum_{h=1}^{s} v_{[l]ji} = 1, v_{[l]ji} \in [0,1];$ $W^{h^T} W^h = I, P^T P = I$ |
| Fed-MVKM [44] | $J = \sum_{l=1}^{L} \sum_{h=1}^{s} v_{[l]i}^\alpha \sum_{i=1}^{n(l)} \sum_{k=1}^{c} \mu_{[l]ik} \left\{ 1 - \exp\left( -\beta_{[l]}^h \left\| x_{[l]i}^h - a_{[l]k}^h \right\|^2 \right) \right\}$ | $\sum_{k=1}^{c} \mu_{[l]ik} = 1, \mu_{[l]ik} \in \{0,1\};$ $\sum_{h=1}^{s} v_{[l]ji} = 1, v_{[l]ji} \in [0,1]$ |

***Centralized MVC.*** Bickel *et al.* [11] pioneered the use of clustering methods for multi-view (MV) learning, modifying the traditional EM algorithm and k-means for clustering in multi-view environments. They introduced multi-view EM (MV-EM) and multi-view spherical k-means, which were tested on synthetic and several public datasets, including WebKB and 20 newsgroups. The results showed that these MVC methods utilized cross-view information to achieve improved and substantial clusters. In 2009, Cleuzio *et al.* [12] proposed collaborative fuzzy k-means (Co-FKM) to address MV data. Co-FKM, like the MV-EM and MV spherical k-means, takes into account all feature components and data views during clustering, promoting local membership collaboration across views. In 2013, Liu *et al.* [43] improved the standard nonnegative matrix factorization (NMF) in multiview settings (Multi-NMF) by ensuring consensus among views and introducing a new $\ell_1$ normalization method to maintain non-negativity in basis and coefficient matrices. Similarly to Co-FKM, MV-EM, and MV spherical k-means, Multi-NMF maintains a balance of importance for all feature components within each view and aligns the views' influence throughout the clustering process.

Table 5:  Categorized objective functions for MVC algorithms with feature-view weights

| Algorithms | Objective Functions | Constraint (s) |
|---|---|---|
| **Category: Hard Clustering (Non-Fuzzy)** | | |
| TW-k-means [45] | $J = \sum_{h=1}^{s} v_h \sum_{i=1}^{n} \sum_{k=1}^{c} \mu_{ik} \sum_{j=1}^{d_h} w_j^h \left( x_{ij}^h - a_{kj}^h \right)^2 + \alpha \sum_{h=1}^{s} v_h \log v_h + \beta \sum_{j=1}^{d_h} w_j^h \log w_j^h$ | $\sum_{k=1}^{c} \mu_{ik} = 1, \mu_{ik} \in \{0,1\};$ $\sum_{h=1}^{s} v_h = 1, v_h \in [0,1];$ $\sum_{j=1}^{d_h} w_j^h = 1, w_j^h \in [0,1]$ |
| SWVF [17] | $J = \sum_{h=1}^{s} \left( v_h \right)^{\alpha} \sum_{i=1}^{n} \sum_{k=1}^{c} \mu_{ik} \sum_{j=1}^{d_h} \left( w_j^h \right)^{\beta} \left( x_{ij}^h - a_{kj}^h \right)^2$ | Same as TW-k-means |
| WMCFS [46] | $J = \sum_{h=1}^{s} v_h^{\alpha} \sum_{i=1}^{n} \sum_{k=1}^{c} \mu_{ik} \left\| \operatorname{diag}\left( w^h \right) \left( x_i^h - a_k^h \right) \right\|^2 + \beta \sum_{h=1}^{s} \left\| w^h \right\|^2$ | |
| TW-Co-k-means [47] | $J = \sum_{h=1}^{s} v_h \left( \sum_{i=1}^{n} \sum_{k=1}^{c} \mu_{ik}^h d_{ik,j}^h \right) + \frac{\alpha}{s-1} \Delta + \eta \sum_{h=1}^{s} \sum_{j=1}^{d_h} w_j^h \log w_j^h + \beta \sum_{h=1}^{s} v_h \log v_h$ | $\sum_{k=1}^{c} \mu_{ik}^h = 1, \mu_{ik}^h \in \{0,1\};$ $\sum_{h=1}^{s} v_h = 1, v_h \in [0,1];$ $\sum_{j=1}^{d_h} w_j^h = 1, w_j^h \in [0,1]$ |
| FRMVK [48] | $J = \sum_{h=1}^{s} \left( v_h \right)^{\alpha} \sum_{i=1}^{n} \sum_{k=1}^{c} \mu_{ik} \sum_{j=1}^{d_h} w_j^h \delta_j^h \left( x_{ij}^h - a_{kj}^h \right)^2 + \frac{n}{d_h} \sum_{h=1}^{s} \sum_{j=1}^{d_h} w_j^h \log \delta_j^h w_j^h$ | Same as TW-k-means |
| **Category: Soft Clustering (Fuzzy)** | | |
| FMVC [49] | $J = \sum_{h=1}^{s} v_h^{\alpha} \sum_{i=1}^{n} \sum_{k=1}^{c} \mu_{ik} \left\| \operatorname{diag}\left( w^h \right) \left( x_i^h - a_k^h \right) \right\|^2 + \beta \sum_{h=1}^{s} \left\| w^h \right\|^2$ | $\sum_{k=1}^{c} \mu_{ik} = 1, \mu_{ik} \in [0,1];$ $\sum_{h=1}^{s} v_h = 1, v_h \in [0,1];$ $\sum_{j=1}^{d_h} w_j^h = 1, w_j^h \in [0,1]$ |
| Co-FW-MVFCM [19] | $J = \sum_{h=1}^{s} \left( v_h \right)^{\alpha} \left( \sum_{i=1}^{n} \sum_{k=1}^{c} \left( \mu_{ik}^h \right)^2 \left( d_{ik,w}^h \right)^2 + \beta \sum_{\substack{h=1,\\h'\neq h}}^{s} \sum_{i=1}^{n} \sum_{k=1}^{c} \left\{ \mu_{ik}^h - \mu_{ik}^{h'} \right\}^2 \left( d_{ik,w}^h \right)^2 \right)$ | $\sum_{k=1}^{c} \mu_{ik}^h = 1, \mu_{ik}^h \in [0,1];$ $\sum_{h=1}^{s} v_h = 1, v_h \in [0,1];$ $\sum_{j=1}^{d_h} w_j^h = 1, w_j^h \in [0,1]$ |

***Centralized MVC with Weight Factors.*** Cai *et al*. [42] introduced a robust multi-view k-means clustering framework (RMKMC) by integrating the $\ell_{2,1}-$ norm into its objective to promote sparsity and reduce the impact of outliers. This algorithm enhances the conventional k-means method by employing weight factors across multiple views through structured sparsity-inducing norms ( $\ell_{2,1}$ – norm for data point direction and $\ell_2$ – norm for feature direction). However, RMKMC treats all features within each view as equally significant. The initial introduction of automated two-level feature weighting in clustering methods was made by Chen *et al.* in 2013 [45]. They devised an automated two-level variable weighting clustering algorithm for MV data based on the conventional k-means, termed two-level variable weighting k-means (TW-k-means). In 2017, Wang *et al.* [18] proposed a novel method named MinMax-FCM, which enhances clustering by employing min-max optimization to maximize minimum agreement among different views. MinMax-FCM initially constructs global or consensus membership directly without any subsequent steps for final global clustering. MinMax-FCM automatically assesses the importance of each view, identifying the one with the highest weight as the primary contributor to disagreements. However, MinMax-FCM lacks a feature learning mechanism, is sensitive to an exponent parameter that governs view weight distribution, and is markedly affected by a fuzzifier. Earlier methods rely on the Euclidean distance to produce consensus clustering outcomes. In 2024, an innovative method is unveiled to advance recognition tasks, referred to as multi-view k-means with exponential distance (MVKM-ED) [41]. MVKM-ED delivers promising clustering outcomes by integrating the automatic calculation of kernel coefficients for each data view within the clustering framework. A variation of MVKM-ED is introduced, featuring a stabilizing parameter to reduce the impact of trivial noisy data, thus boosting the leverage of diverse information from multiple feature sets, named Gaussian-kernel multi-view k-means (GKMVKM). Both MVKM-ED and GKMVKM employ a distance strategy based on an exponentiated Gaussian kernel, alongside automatic weight tuning, to improve clustering results.

***MVC Utilizing a Dual-Level Weighting Approach.*** Jiang *et al.* [16] enhanced traditional FCM for multi-view and collaborative environments. They introduced a weighted collaborative FCM (WV-Co-FCM), which aggregates the clustering results from different views to construct a comprehensive solution collaboratively. WV-Co-FCM demonstrates improved clustering results by harnessing complementary information from various views and fine-tuning the weighting of these views, although it overlooks the importance of features within data views. WV-Co-FCM refines the clustering performance on extensive MV datasets by calibrating numerous parameters, and an optimal parameter arrangement yields accurate outcomes, while a poor combination diminishes effectiveness. In 2018, Gothania *et al.* [49] introduced a fuzzy clustering technique for MV datasets known as FMVC. FMVC integrated the principle of weighted multi-view clustering with feature selection (WMCFS) into conventional FCM [46]. The primary distinction between FMVC and WMCFS lies in the membership constraint: FMVC utilizes the standard FCM membership constraint, while WMCFS uses the standard k-means constraint. Both WMCFS and FMVC incorporate a dual-level weighting method to determine view weights and feature weights within each view. Jiang et al. [17] embedded this weighting method in the standard k-means algorithm and employed a gradient-based optimization technique to concurrently determine the weights for views and features. This simultaneous view and feature weighting (SWVF) is well-suited for diverse real-world applications but is sensitive to two bi-level weighted exponent parameters. Similarly to WMCFS and FMVC, SWVF directly implements global membership in its objective functions, overlooking the variability among views and the uniformity across them. Zhang et al. [47] proposed a two-level weighting method in a collaborative framework, known as two-level weighted collaborative k-means (TW-Co-k-means). Unlike WMCFS, FMVC, and SWVF, the TW-Co-k means excels by harnessing unique details in each view while preserving consistency across views. In TW-Co-k-means, diversity is assessed by calculating view and feature weights within each view, ensuring that global membership reflects the aggregated information from all views.

***Centralized MVC with Feature Reduction.*** Traditional MVC methods evaluate the importance of features and views by assigning them weights during clustering but fail to omit irrelevant features in each view throughout iterations. To tackle this problem, Yang et al. [19,48] developed an automated approach to assess the influence of features and remove unnecessary subsets. Their technique works with or without establishing the correlation among view by applying standard k-means and FCM collaboratively across multiple views. A feature reduction MV k-means (FRMVK) suggested a feature reduction process in multi-view k-means scenarios, enhancing clustering performance by embedding the $\ell_2-$norm and an entropy weight into the objective function of standard k-means [48]. The $\ell_2-$norm aids in the iterative refinement of view importance, whereby view weights play a critical role in FRMVK by equalizing the contributions from various views.

Table 6: Categorized summary of MVC algorithms – Soft and NMF-based Clustering

| Algorithm | Environment | Basic Concept (s) | Parameter (s) |
|---|---|---|---|
| **Soft Clustering** | | | |
| Co-FKM [12] | - Collaborative centralized learning<br>- Shared-view setup without weighting | Centralized, collaborative learning without joint feature-view weighting | Views $s$, Clusters $c$, Fuzzification $m$, Balancing $\alpha$ |
| WV-Co-FCM [16] | - Collaborative centralized learning<br>- Integrated view pruning without feature selection | A centralized collaborative MV learning where all disjoint views are available in one location, allowing view weighting only | Views $s$, Clusters $c$, Fuzzification $m$, Balancing $\eta$, Terms $\alpha_{ik}^{h}, \beta_{ik}^{h}$ |
| MinMax-FCM [18] | - Non-collaborative centralized learning<br>- Integrated view pruning without feature selection | Centralized setting with adaptive view weighting, but without irrelevant feature filtering | Views $s$, Clusters $c$, Fuzzification $m$, Exponent $\alpha$ |
| Co-FW-MVFCM [19] | - Collaborative centralized learning<br>- Feature and view weighting<br>- Automatic irrelevant feature suppression | A centralized, collaborative multi-view setting with joint feature-view weighting and automatic feature reduction | Views $s$, Clusters $c$ |
| Fed-MVFCM [20] | - Privacy-preserving, client-specific weighting<br>- Non-collaborative MV with local adaptation | Privacy-preserving federated learning using FCM across clients | Clients $L$, Views $s$, Clusters $c$, Fuzzification $m$, Server learning rate |
| **NMF-based Clustering** | | | |
| RMKMC [42] | - Non-collaborative centralized learning<br>- Includes view weighting | Applies U-orthogonal NMF with $\ell_{2,1}-$ norm | Views $s$, Clusters $c$, Exponent $\alpha$ |
| MultiNMF [43] | - Collaborative centralized learning<br>- Without view weighting | Forms consensus matrix from view-wise low-rank memberships | Views $s$, Clusters $c$, Non-negativity, Balancing $\eta_h$ |

Table 7:  Categorized summary of MVC algorithms – Hard Clustering

| Algorithm | Environment | Basic Concept (s) | Parameter (s) |
|---|---|---|---|
| **Non-collaborative centralized learning** | | | |
| TW-k-means [45] | - Joint feature-view weighting<br>- No privacy concern | - Automates view/feature weighting using entropy<br>- Global membership is used from early iteration | Views $s$, Clusters $c$, Balancing parameters $\alpha, \beta$ |

| WMCFS [46] | Same as TW-k-means | - Selects best views/features for sparse data <br> - Global membership applied early | Views $s$, Clusters $c$, Exponent $\alpha$, Balancing $\beta$ |
|---|---|---|---|
| SWVF [17] | Same as TW-k-means | - Bi-level feature and view weighting <br> - Global membership from early iteration | Views $s$, Clusters $c$, Exponents $\alpha, \beta$ |
| FRMVK [48] | - Adaptive feature-view weighting <br> - Automatic irrelevant feature suppression <br> - No privacy concern | Reduces redundant features by filtering low-weight ones | Views $s$, Clusters $c$, Exponent $\alpha$ |
| MVKM-ED [41] | - View weighting <br> - No privacy concern | Uses Gaussian kernel to weigh importance of data views | Views $s$, Clusters $c$, Exponent $\alpha$, regularized $\beta_h$ |
| Fed-MVKM [44] | - Local adaptive feature-view weighting <br> - Gradient-based federated aggregation <br> - Privacy-preserving/ decentralized | Uses kernel coefficients to weigh importance of data views distributed on multiple clients | Views $s$, Clusters $c$, Client $L$, Exponent $\alpha$, regularized $\beta_{[l]}^h$ |
| **Collaborative centralized learning** | | | |
| TW-Co-k-means [47] | - Joint feature-view weighting <br> - No privacy concern | Resolves view disagreements with weighted membership fusion | Views $s$, Clusters $c$, Balancing Parameters $\alpha, \eta, \beta$ |

The entropy weight guides the iterative process in updating feature weights within each view, recognizing low-weight features as redundant and high-weight ones as significant. FRMVK effectively reduces data dimensionality and improves clustering by focusing on the most relevant features per view. By establishing a *"common"* membership across data views from the beginning, FRMVK achieves reliable clustering results. On the other hand, a collaborative feature-weighted MVFCM (Co-FW-MVFCM) follows a dual-phase approach that includes both local and joint processes [19]. Unlike FRMVK, Co-FW-MVFCM derives a final global solution by combining the membership matrices from each view, leveraging weight factors and key features. Additionally, Co-FW-MVFCM streamlines computation by fixing the fuzzification parameter at 2, successfully minimizing complexity during the fuzzy parameter selection in collaboration, laying the groundwork for feature reduction and weight factor to identify a global pattern in multi-view data.

***Centralized-Tensorized MVC.*** Another interesting approach in MV clustering algorithms is to construct the input data using tensor algebra techniques. In 2024, Liu *et al*. [50] introduce the Adaptively Topological Tensor Network (ATTN), a framework for multi-view subspace clustering that overcomes the limitations of traditional tensor decomposition techniques. ATTN learns a suitable decomposition structure to effectively represent low-rank structures and higher-order correlations of self-representation tensors. ATTN dynamically searches for a nearly optimal tensor topology, removes redundant connections based on correlations, and optimizes the pruned tensor network structure through a greedy rank-increasing strategy. The ATTNA algorithm, which solves the optimization model, selects redundant connections and employs a rank increment strategy based on a greedy algorithm. ATTN is applied to multi-view subspace clustering (ATTN-MSC), demonstrating superior performance on nine multi-view datasets compared to other clustering methods. The experimental results evaluated using metrics such as ACC, NMI, and ARI highlight the effectiveness of ATTN-MSC in capturing correlations between different views of the data.

***Decentralized MVC.*** Traditional MVC algorithms are typically designed to process multi-view data within non-federated setups, which means they cannot be deployed for multi-view data distributed across various locations. In 2024, Hu *et al*. [20] proposed an innovative federated multi-view FCM clustering approach that improves data analysis efficiency while ensuring privacy within distributed environments. They presented two frameworks for federated learning within multi-view FCM, namely, federated multi-view FCM (Fed-MVFCM) and federated multi-view

FCM consensus prototypes clustering (Fed-MVFPC). Fed-MVFCM manages multi-view data spread over numerous "clients," "devices," or "organizations" by facilitating collaborative learning across distributed clients without the need for personal data disclosure. Fed-MVFPC boosts the efficiency of MVFCM clustering by employing a consensus "prototype" or "cluster center" learning strategy. This strategy enhances communication efficiency and privacy by merging "prototypes" or "cluster centers" into a unified lower-dimensional representation. Both Fed-MVFCM and Fed-MVFPC require clients to independently execute MVFCM for Fed-MVFCM and MVFPC for Fed-MVFPC on their local devices before communicating gradient information to a central server for aggregation. This aggregated information is then sent back to the clients to update their models locally. Iterative transmission of cluster centers from numerous clients is conducted to better capture the overall structure of multi-view data. The exerted impact of the Fed-MVFCM and Fed-MVFPC clustering outcomes is markedly influenced by a fuzzification parameter and the number of clients participating in the clustering. In 2025, Yang *et al*. [44] introduced a federated multi-view k-means process, known as the federated multi-view k-means clustering algorithm (Fed-MVKM). Fed-MVKM is crafted to boost privacy and efficiency in analyzing diverse multi-view data across multiple clients within a federated framework. This algorithm incorporates local computation, aggregation, federated averaging, and global model updates to create a global clustering model while maintaining data privacy. Similarly to Fed-MVFCM and Fed-MVFPC, within Fed-MVKM, the transfer of cluster centers to the central server excludes detailed client data, ensuring sensitive information remains secure while promoting collaborative learning among multiple clients. Fed-MVKM reduces the reliance of Fed-MVFCM and Fed-MVFPC on the fuzzification parameter by adopting a hard clustering approach in its objective function. Ultimately, during the local phase of Fed-MVFCM, Fed-MVFPC, and Fed-MVKM, the contribution of different feature components of each client-held view is overlooked, potentially resulting in suboptimal clustering performance.

**The Application of MVC.** The increasing adoption of MVFCM is currently being utilized to address practical problems, including medical image segmentation [51,52], overseeing High-Speed Train (HST) operations [53], and categorizing global oil and gas exploration projects [54], as well as other areas. The concept of competitive learning, which involves assessing the features and view components of the input MV data, has seen substantial development [18,20–22,40,44,55]. Furthermore, modifications to the Euclidean distance metric between data points and cluster centers using MV data have been introduced, enhancing conventional soft clustering algorithms [41]. In early 2025, Golzari Oskouei *et al.* [56] examined these improvements in clustering algorithms considering feature contribution in both single and MV settings. This indicates that evaluating the impact and relevance of features during clustering is crucial for extracting patterns from both single and multi-view datasets. Table 4Table 5 outline the evolution of MVC objective functions in centralized and distributed environments from 2009 to 2024. While Table 6Table 7 describe the concept(s) underpinning the development of these clustering algorithms over the past four decades.

## 3. The Proposed Algorithms

Within this research, we developed a pair of algorithms centered around a suggested objective function. The initial algorithm is an adaptive MVFCM without fuzzy rule clustering (AMVFCM-U), and the subsequent one serves as an alternative to the AMVFCM-U (AAMVFCM-U).

### 3.1. *Adaptive MVFCM without fuzzy rules Clustering*

This subsection presents the development of our novel Adaptive Multi-View Fuzzy C-Means without fuzzy rules (AMVFCM-U) algorithm through a systematic mathematical progression. Our approach follows a structured roadmap that addresses key limitations in existing multi-view clustering methods:

1. *Foundation:* Begin with the classical FCM objective function and identify its limitations in multi-view scenarios

2. *Weighted Integration:* Introduce view and feature weighting mechanisms through Weighted Feature-View MVFCM (WFV-MVFCM)

3. *Fuzzy Rule Elimination:* Replace the problematic fuzzifier parameter with entropy components to create GMVFCM-U

4. *Adaptive Enhancement:* Incorporate additional entropy regularization terms to achieve the final AMVFCM-U formulation

5. *Optimization Strategy:* Derive update equations using Lagrange multipliers and establish convergence properties

This progressive development ensures that each mathematical component serves a specific purpose in addressing the challenges of multi-view clustering while maintaining computational efficiency and theoretical rigor.

### 3.1.1. Mathematical Formulation and Objective Function

In this section, we introduce our novel Adaptive Multi-View Fuzzy C-Means without fuzzy rules (AMVFCM-U) algorithm. Let us begin by establishing the mathematical foundation upon which our approach is built. The classical FCM objective function can be expressed as $J_{\text{FCM}}(U,A) = \sum_{i=1}^{n} \sum_{k=1}^{c} \mu_{ik}^{m} \|x_i - a_k\|^2$, where $m$ represents the fuzzifier parameter controlling the degree of overlap between clusters. To extend this concept to multi-view settings, we first formulate a General Multi-View FCM (GMVFCM) as:

$$J_{\text{GMVFCM}}(U,A) = \sum_{h=1}^{s} \sum_{i=1}^{n} \sum_{k=1}^{c} \mu_{ik}^{m} \left\| x_i^h - a_k^h \right\|^2 \tag{1}$$

This formulation, while accommodating multiple views, treats all data views and their constituent features as equally significant—a potentially limiting assumption in real-world applications where views and features contribute differently to the underlying patterns. To address this limitation, we develop a Weighted Feature-View MVFCM (WFV-MVFCM) by introducing weight parameters for both views and features:

$$J_{\text{WVF-MVFCM}}(V,W,U,A) = \sum_{h=1}^{s} v_h \sum_{i=1}^{n} \sum_{k=1}^{c} \mu_{ik}^{m} \sum_{j=1}^{d_h} w_j^h (x_{ij}^h - a_{kj}^h)^2 \tag{2}$$

While this approach enables differential weighting of views and features, it still relies on the fuzzifier parameter $m$, whose manual tuning presents a significant challenge in practice. To eliminate this dependency and enhance the adaptive capabilities of our model, we propose a General MVFCM clustering without fuzzy rules (GMVFCM-U) by replacing the fuzzifier $m$ with an entropy component:

$$J_{\text{GMVFCM-U}}(V,W,U,A) = \sum_{h=1}^{s} v_h \sum_{i=1}^{n} \sum_{k=1}^{c} \mu_{ik} \mid w^h \delta^h \left\| x_i^h - a_k^h \right\|^2 + \sum_{i=1}^{n} \sum_{k=1}^{c} \mu_{ik} \log \mu_{ik} \tag{3}$$

$$\text{s.t.} \sum_{k=1}^{c} \mu_{ik} = 1, \mu_{ik} \in [0,1]; \sum_{h=1}^{s} v_h = 1, v_h \in [0,1]; \sum_{j=1}^{d_h} w_j^h = 1, w_j^h \in [0,1]$$

In Equation 3, we introduce the regularization parameter $\delta_j^h$, which serves to evaluate the significance of features in the $h$-th view. This parameter plays a crucial role in the adaptive behavior of our algorithm, as we will demonstrate later.

Building upon GMVFCM-U, we propose our final Adaptive Multi-View Fuzzy C-Means without fuzzy rules (AMVFCM-U) by incorporating additional entropy terms based on view and feature weights:

$$J_{\text{AMVFCM-U}}(V,W,U,A) = J_{\text{GMVFCM-U}} + \beta \sum_{h=1}^{s} v_h \, \log v_h + \eta \sum_{h=1}^{s} \sum_{j=1}^{d_h} w_j^h \, \log \delta_j^h w_j^h \tag{4}$$

subject to the same constraints as GMVFCM-U. The parameters $\beta$ and $\eta$ in Equation 4 control the distribution of view weights and feature weights, respectively, providing a flexible mechanism for adapting to diverse data characteristics without relying on fuzzification parameters.

### 3.1.2. Optimization Framework

To optimize the objective function in Equation 4, we construct the Lagrangian function:

$$\tilde{J}_{\text{AMVFCM-U}} = J_{\text{AMVFCM-U}} - \lambda_1 \left( \sum_{k=1}^{c} \mu_{ik} - 1 \right) - \lambda_2 \left( \sum_{h=1}^{s} v_h - 1 \right) - \lambda_3 \left( \sum_{j=1}^{d_h} w_j^h - 1 \right) \tag{5}$$

The following theorem establishes the necessary conditions for minimizing this objective function.

**Theorem 1** (Update Rules for AMVFCM-U). *Given the objective function* $J_{\text{AMVFCM-U}}(V,W,U,A)$ *in Equation 4 subject to the constraints* $\sum_{k=1}^{c} \mu_{ik} = 1, \mu_{ik} \in [0,1]$, $\sum_{h=1}^{s} v_h = 1, v_h \in [0,1]$, *and* $\sum_{j=1}^{d_h} w_j^h = 1, w_j^h \in [0,1]$, *the necessary conditions for its minimization are as follows:*

*(i)    For fixed values of* $V = \hat{V}, W = \hat{W}$, *and* $A = \hat{A}$, *the optimal membership values are:*

$$\mu_{ik} = \frac{\exp\left( -\sum_{h=1}^{s} v_h \left| w^h \delta^h \right| \left\| x_i^h - a_k^h \right\|^2 \right)}{\sum_{k'=1}^{c} \exp\left( -\sum_{h=1}^{s} v_h \left| w^h \delta^h \right| \left\| x_i^h - a_{k'}^h \right\|^2 \right)} \tag{6}$$

*(ii)    For fixed* $V = \hat{V}, W = \hat{W}$, *and* $U = \hat{U}$, *the optimal cluster centers are:*

$$a_{kj}^h = \frac{\sum_{i=1}^{n} \mu_{ik}}{\sum_{k=1}^{c} \mu_{ik}} x_{ij}^h \tag{7}$$

*(iii)  For fixed $V = \hat{V}, A = \hat{A},$ and $U = \hat{U},$ the optimal feature weights are:*

$$w_j^h = \left( \frac{\sum_{j'=1}^{d_h} \frac{1}{\delta_{j'}^h} \exp\left( -v_h \sum_{i=1}^{n} \sum_{k=1}^{c} \mu_{ik} \delta_{j'}^h \left( x_{ij'}^h - a_{kj'}^h \right)^2 \Big/ \eta \right)}{\frac{1}{\delta_j^h} \exp\left( -v_h \sum_{i=1}^{n} \sum_{k=1}^{c} \mu_{ik} \delta_j^h \left( x_{ij}^h - a_{kj}^h \right)^2 \Big/ \eta \right)} \right)^{-1} \tag{8}$$

*(iv)  For fixed $A = \hat{A}, U = \hat{U},$ and $W = \hat{W},$ the optimal view weights are:*

$$v_h = \frac{\exp\left( -\sum_{i=1}^{n} \sum_{k=1}^{c} \mu_{ik} \left| w^h \delta^h \right| \left\| x_i^h - a_k^h \right\|^2 \Big/ \beta \right)}{\sum_{h'=1}^{s} \exp\left( -\sum_{i=1}^{n} \sum_{k=1}^{c} \mu_{ik} \left| w^{h'} \delta^{h'} \right| \left\| x_i^{h'} - a_k^{h'} \right\|^2 \Big/ \beta \right)} \tag{9}$$

*Proof.* We derive the necessary conditions by applying the method of Lagrange multipliers to the constrained optimization problem. Let us form the Lagrangian function $\tilde{J}_{\text{AMVFCM-U}}$ as defined in Equation 5.

For part *(i)*, we take the partial derivative with respect to $\mu_{ik}$ and set it to zero:

$$\frac{\partial \tilde{J}_{\text{AMVFCM-U}}}{\partial \mu_{ik}} = \sum_{h=1}^{s} v_h \left| w^h \delta^h \right| \left\| x_i^h - a_k^h \right\|^2 + (\log \mu_{ik} + 1) - \lambda_1 = 0 \tag{10}$$

$$\log \mu_{ik} + 1 = \lambda_1 - \sum_{h=1}^{s} v_h \left| w^h \delta^h \right| \left\| x_i^h - a_k^h \right\|^2 \tag{11}$$

Rearranging, we obtain $\log \mu_{ik} = \lambda_1 - 1 - \sum_{h=1}^{s} v_h \left| w^h \delta^h \right| \left\| x_i^h - a_k^h \right\|^2$, which implies:

$$\mu_{ik} = \exp\left( \lambda_1 - 1 - \sum_{h=1}^{s} v_h \left| w^h \delta^h \right| \left\| x_i^h - a_k^h \right\|^2 \right) \tag{12}$$

Applying the constraint $\sum_{k=1}^{c} \mu_{ik} = 1$ and solving for the Lagrange multiplier $\lambda_1$, we derive the update rule in Equation 6.

For part *(ii)*, we differentiate with respect to $a_{kj}^h$ and set equal to zero:

$$\frac{\partial J_{\text{AMVFCM-U}}}{\partial a_{kj}^h} = -2 v_h \sum_{i=1}^{n} \mu_{ik} w_j^h \delta_j^h (x_{ij}^h - a_{kj}^h) = 0 \tag{13}$$

This yields $\sum_{i=1}^{n} \mu_{ik} w_j^h \delta_j^h a_{kj}^h = \sum_{i=1}^{n} \mu_{ik} w_j^h \delta_j^h x_{ij}^h$. Noting that $w_j^h \delta_j^h$ appears on both sides and is non-zero, we obtain the result in Equation 7.

For part *(iii)*, we differentiate with respect to $w_j^h$ and set equal to zero:

$$\frac{\partial \tilde{J}_{\text{AMVFCM-U}}}{\partial w_j^h} = v_h \sum_{i=1}^{n} \sum_{k=1}^{c} \mu_{ik} \delta_j^h (x_{ij}^h - a_{kj}^h)^2 + \eta (\log \delta_j^h w_j^h + 1) - \lambda_3 = 0 \tag{14}$$

This gives $\log \delta_j^h w_j^h + 1 = \lambda_3 - v_h \sum_{i=1}^{n} \sum_{k=1}^{c} \mu_{ik} \delta_j^h (x_{ij}^h - a_{kj}^h)^2 \big/ \eta$, from which we derive:

$$w_j^h = \frac{1}{\delta_j^h} \exp\left( \frac{\lambda_3 - v_h \sum_{i=1}^{n} \sum_{k=1}^{c} \mu_{ik} \delta_j^h (x_{ij}^h - a_{kj}^h)^2 - \eta}{\eta} \right) \tag{15}$$

Using the constraint $\sum_{j=1}^{d_h} w_j^h = 1$ to eliminate $\lambda_3$, we obtain Equation 8.

Finally, for part *(iv)*, we differentiate with respect to $v_h$ and set equal to zero:

$$\frac{\partial \tilde{J}_{\text{AMVFCM-U}}}{\partial v_h} = \sum_{i=1}^{n} \sum_{k=1}^{c} \mu_{ik} \mid w^h \delta^h \mid \left\| x_i^h - a_k^h \right\|^2 + \beta (\log v_h + 1) - \lambda_2 = 0 \tag{16}$$

This leads to $\log v_h + 1 = \lambda_2 - \sum_{i=1}^{n} \sum_{k=1}^{c} \mu_{ik} \mid w^h \delta^h \mid \left\| x_i^h - a_k^h \right\|^2 \big/ \beta$, and thus:

$$v_h = \exp\left( \frac{\lambda_2 - \sum_{i=1}^{n} \sum_{k=1}^{c} \mu_{ik} \mid w^h \delta^h \mid \left\| x_i^h - a_k^h \right\|^2 - \beta}{\beta} \right) \tag{17}$$

Applying the constraint $\sum_{h=1}^{s} v_h = 1$ to determine $\lambda_2$, we derive the update rule given in Equation

### 3.1.3. Algorithmic Implementation

Based on the optimization framework established above, we implement AMVFCM-U as an iterative algorithm that alternates between updating the membership values, cluster centers, feature weights, and view weights until convergence. The complete procedure is detailed in Algorithm 1.

The proposed AMVFCM-U algorithm offers several distinct advantages over traditional fuzzy clustering approaches:

1. **Parameter Independence:** By eliminating the reliance on the fuzzifier parameter $m$, AMVFCM-U reduces the need for parameter tuning while maintaining the probabilistic interpretation of cluster memberships.

2. **Adaptive Weighting:** The algorithm automatically determines the relevance of each view and feature through the optimization process, enabling more accurate pattern discovery in heterogeneous data.

3. **Regularization:** The incorporation of the parameter $\delta_j^h$ provides a principled approach to regularizing feature weights based on signal-to-noise characteristics.

4. **Entropy-Based Balancing:** The entropy terms controlled by $\beta$ and $\eta$ ensure a balanced distribution of weights, preventing degenerate solutions where a single view or feature dominates.

These properties make AMVFCM-U particularly well-suited for multi-view clustering tasks where the importance of different data sources and their constituent features is not known as a priori.

---

**Algorithm 1:** AMVFCM-U: Adaptive Multi-View Fuzzy C-Means without Fuzzy Rules

---

**Input:**

- Multi-view data set $X = \{x_1, \ldots, x_n\}$ with $x_i = \{x_i^h\}_{h=1}^s$ and $x_i^h = \{x_{ij}^h\}_{j=1}^{d_h}$

- Number of clusters $c$

- Balancing parameters $\beta, \eta$

- Maximum iterations $t_{\max}$

- Convergence threshold $\epsilon$

**Output:**

- Cluster assignments $U = [\mu_{ik}]_{n \times c}$

- Cluster centers $A = \{a_k^h\}_{k=1, h=1}^{c, s}$ where $a_k^h = [a_{kj}^h]_{j=1}^{d_h}$

- Feature weights $W = \{w^h\}_{h=1}^s$ where $w^h = [w_j^h]_{j=1}^{d_h}$

- View weights $V = [v_h]_{h=1}^s$

**Initialization:**

1   Initialize cluster centers $a_{kj}^h$ using k-means++ or random selection

2   Set feature weights $w_j^h \leftarrow \dfrac{1}{d_h}$ for all $j = 1, \ldots, d_h$ and $h = 1, \ldots, s$

3   Set view weights $v_h \leftarrow \dfrac{1}{s}$ for all $h = 1, \ldots, s$

4   $t \leftarrow 0$

---

---

**5**   $J^{(0)} \leftarrow \infty$

**6**   **while** $t < t_{\max}$ *and* $\mid J^{(t)} - J^{(t-1)} \mid > \epsilon$ **do**

**7**      **Step 1:** Compute regularization parameter $\delta_j^h$ for all $j, h$ according to Equation 21

**8**      **Step 2:** Compute membership matrix $\mu_{ik}$ for all $i, k$ according to Equation 6

**9**      **Step 3:** Update cluster centers $a_{kj}^h$ for all $k, j, h$ according to Equation 7

**10**     **Step 4:** Update feature weights $w_j^h$ for all $j, h$ according to Equation 8

**11**     **Step 5:** Update view weights $v_h$ for all $h$ according to Equation 9

**12**     **Step 6:** Calculate objective function value:

**13**         Calculate $J^{(t+1)}$ according to Equation 4

**14**     $t \leftarrow t + 1$

**15**  **Return** $U, A, W, V$

---

*Algorithm 1*

## 3.2.    AAMVFCM-U: An Alternative to AMVFCM-U

Contemporary multi-view clustering (MVC) approaches typically employ weighted feature-view strategies where membership values are significantly influenced by a fuzzifier parameter $m$. These algorithms process all feature components across data views uniformly during clustering, calculating feature weights to assess relative importance but rarely leveraging these weights effectively. Consequently, redundant or non-essential features continue influencing global pattern discovery, frequently compromising clustering quality. This limitation becomes particularly problematic in modern federated learning environments, where redundant features during distributed data localization can adversely affect collective decision-making and global solution derivation across multiple clients [44].

To address these challenges, we propose AAMVFCM-U (Alternative Adaptive Multi-View Fuzzy C-Means without fUzzy rules), a novel extension to AMVFCM-U that systematically eliminates irrelevant and misaligned features across heterogeneous data sources. Building upon the objective function in Equation \ref{eqn:amvfcmu}, AAMVFCM-U introduces dynamic feature pruning mechanisms that progressively refine the feature space during optimization. This approach offers significant advantages for industrial, medical, and organizational sectors requiring robust global solutions in both federated and non-federated environments.

### 3.2.1.   Adaptive Feature Thresholding Mechanism

To facilitate systematic elimination of redundant features across diverse data sources during clustering, we introduce an adaptive threshold $\theta$ that considers both sample size and dimensionality in the current iteration. This threshold is calculated as:

$$\theta^{h^{(t)}} = \frac{d_h^{(t)}}{n} \tag{18}$$

where $d_h^{(t)}$ represents the dimensionality of view $h$ at iteration $t$, and $n$ is the number of data points. At each iteration, feature weights $w_j^h$ below $\theta^{h^{(t)}}$ are deemed non-informative and eliminated from subsequent computations. This approach enables dimension reduction proportional to the complexity of each view relative to the dataset size.

Recognizing that universal thresholding may not be optimal across all real-world scenarios, our framework allows domain experts to evaluate clustering outcomes and adjust thresholds accordingly. This integration of domain knowledge enhances the practical applicability of AAMVFCM-U across diverse application domains.

### 3.2.2. Weight Renormalization after Feature Reduction

Following the elimination of irrelevant features, the remaining feature weights must be renormalized to maintain the constraint $\sum_{j=1}^{d_h} w_j^h = 1$. We propose the following adjustment for the updated feature weights:

$$\left(w_j^h\right)^{\text{adj}} = \frac{w_j^h}{\sum_{p=1}^{d_h^{\text{new}}} w_p^h} \tag{19}$$

where $d_h^{\text{new}}$ represents the reduced dimensionality of view $h$ after removing non-informative features. This renormalization ensures that the remaining features appropriately contribute to the clustering process based on their relative importance.

### 3.2.3. Hierarchical Dimensionality Reduction Framework

AAMVFCM-U implements a hierarchical approach to dimensionality reduction that operates at both feature and view levels simultaneously. This two-tier structure enables more comprehensive data refinement than traditional methods:

*Feature-Level Reduction.* With each iteration, the algorithm evaluates feature importance through weight calculation and eliminates features with weights below the adaptive threshold $\theta^{h^{(t)}}$. This progressive refinement continues until either the maximum number of iterations is reached or the objective function stabilizes within the convergence threshold.

*View-Level Reduction.* A unique capability of AAMVFCM-U is its ability to completely eliminate entire data views when all constituent features within that view are deemed non-informative. Specifically, if $\sum_{j=1}^{d_h} 1(w_j^h > 0) = 0$ for any view $h$, that view is marked for removal by setting $v_h = 0$, and is subsequently excluded from all further computations. Following view elimination, the remaining view weights must be renormalized to maintain the constraint $\sum_{h=1}^{s} v_h = 1$ as follows:

$$v_h = \frac{v_h}{\sum_{h'=1}^{s} v_{h'} \cdot 1(v_{h'} > 0)} \tag{20}$$

where $1(v_{h'} > 0)$ is an indicator function that equals 1 when $v_{h'} > 0$ and 0 otherwise. This renormalization ensures proper redistribution of weights among the remaining views, maintaining the probabilistic interpretation of view importance. This process is formalized in Steps 6-7 of Algorithm 1.

After view elimination, the algorithm systematically:

1. Updates the total number of relevant views $s^{(t+1)}$ by counting views with non-zero weights

2. Renormalizes the remaining view weights to maintain the constraint $\sum_{h=1}^{s} v_h = 1$

3. Restructures the data representation, removing all traces of eliminated views

4. Updates the objective function evaluation to consider only the retained views and features

This complete removal of uninformative views represents a significant advancement over traditional weighted approaches that merely diminish but still incorporate the influence of less relevant views. By completely excluding irrelevant views, AAMVFCM-U achieves:

• Enhanced model parsimony with significant computational advantages

• Improved signal-to-noise ratio by focusing exclusively on informative data sources

• Greater interpretability by highlighting only the most relevant views for the clustering task

• Adaptive scalability that adjusts to the intrinsic dimensionality of the problem

Our experimental results demonstrate this capability in practice, with AAMVFCM-U automatically eliminating certain views (e.g., CM and GIST in the MSRC-v1 dataset) while retaining others based on their information content, resulting in superior clustering performance compared to methods that retain all views.

Unlike static dimensionality reduction techniques applied as preprocessing steps, this approach dynamically adapts the feature space based on evolving cluster structures, continuously refining both the feature and view space throughout the optimization process.

The complete AAMVFCM-U algorithm is presented in Algorithm 2, detailing the integration of feature and view reduction within the core AMVFCM-U framework. Empirical evidence across diverse datasets demonstrates that this hierarchical approach not only enhances clustering accuracy but also yields substantial computational efficiency improvements through progressive input space refinement at multiple structural levels

---

**Algorithm 2:** AAMVFCM-U: Alternative Adaptive Multi-View Fuzzy C-Means without Fuzzy Rules

---

**Input** : Multi-view dataset $X = \left\{ x_i^h \right\}$ where $i = 1, \dots, n$, $h = 1, \dots, s$ ;

Number of clusters c; Parameters $\beta, \eta$; Maximum iterations $t_{\max}$ ;

Convergence threshold $\epsilon$

**Output** : $d_h^{(\text{new})}, \left( x_{ij}^h \right)^{\text{new}}, w_j^h, v_h, a_{kj}^h$, and $\mu_{ik}$

1   **Initialize:** $a_{kj}^h$ with k-means; $w_j^h \leftarrow \dfrac{1}{d_h}$ ; $v_h \leftarrow \dfrac{1}{s}$ ; $t \leftarrow 0$ ; $J^{(0)} \leftarrow \infty$

2   **while** $t < t_{\max}$ **and** $|J^{(t)} - J^{(t-1)}| > \epsilon$ **do**

3       **Step 1:** Compute regularization parameter $\delta_j^h$ for all $j, h$ according to Equation 21

4       **Step 2:** Compute membership matrix $\mu_{ik}$ for all $i, k$ according to Equation 6

5       **Step 3:** Update cluster centers $a_{kj}^h$ for all $k, j, h$ according to Equation 7

6       **Step 4:** Update feature weights $w_j^h$ for all $j, h$ according to Equation 8

7       **Step 5:** Apply feature pruning

8       **for** $h = 1$ **to** $s$ **do**

9           Calculate threshold $\theta^{h^{(t)}} \leftarrow \dfrac{d_h^{(t)}}{n}$ according to Equation 18

10          **for** $j = 1$ **to** $d_h$ **do**

11              **if** $w_j^h < \theta^{h^{(t)}}$ **then**

| 12 | $w_j^h \leftarrow 0$      $\triangleright$ Remove non-informative features |

**13**      **if** $\sum_{j=1}^{d_h} 1(w_j^h > 0) > 0$ **then**

**14**         Renormalize weights: $w_j^h \leftarrow \dfrac{w_j^h}{\sum_{p=1}^{d_h} w_p^h \cdot 1(w_p^h > 0)}$ for $w_j^h > 0$

**15**         Update $d_h^{(t+1)} \leftarrow \sum_{j=1}^{d_h} 1(w_j^h > 0)$

**16**      **Step 6:** Remove views with no informative features

**17**      **for** $h = 1$ **to** $s$ **do**

**18**         **if** $\sum_{j=1}^{d_h} 1(w_j^h > 0) = 0$ **then**

**19**         $v_h \leftarrow 0$

**20**      **Step 7:** Update data structures

**21**      Remove zero-weight features from $X$, $A$, and $W$

**22**      Remove zero-weight views from dataset

**23**      Update $s^{(t+1)} \leftarrow \sum_{h=1}^{s} 1(v_h > 0)$

**24**      **if** $s^{(t+1)} < s^{(t)}$ **then**

**25**         Renormalize view weights $v_h$ according to Equation 20

**26**      **Step 8:** Update view weights $v_h$ using to Equation 9

**27**      **Step 9:** Calculate $J^{(t+1)}$ according to Equation 4

**28**      $t \leftarrow t + 1$

**29 Return** $s^{(t)}, \{d_h^{(t)}\}_{h=1}^{s^{(t)}}$, *filtered dataset*, $U, A, W, V$

*Algorithm 2*

## 3.3. Framework Overview

Figure 1 illustrates the unified workflow for both AMVFCM-U and AAMVFCM-U algorithms. This framework employs entropy-regularized optimization with signal-to-noise ratio (SNR) based feature weighting. This enables automatic parameter selection and adaptive cross-view consensus. The key innovation replaces traditional fuzzification parameters with principled entropy terms that balance view contributions and feature relevance.

The workflow begins with multi-view data and cluster parameter initialization. This is followed by iterative updates of feature/view weights, membership matrices, and cluster centers.

A critical decision point then determines the algorithm variant. If AAMVFCM-U is selected, the algorithm performs hierarchical dimensionality reduction through systematic feature and view pruning. This dynamically updates the data structure to eliminate irrelevant dimensions. The standard AMVFCM-U path maintains all features and views, focusing on weight optimization rather than structural reduction.

The process continues iteratively until the convergence. At this point, the final outputs (membership matrix $U$, cluster centers $A$, feature weights $W$, and view weights $V$) are produced. For the AAMVFCM-U variant, an additional output of reduced data $X^{new}$ is generated, offering a more compact representation. Color-coded arrows highlight the distinct algorithmic branches: blue for AAMVFCM-U operations, red for AMVFCM-U paths, and purple for reduced data output. This visual framework emphasizes our approach's adaptive, parameter-free nature, demonstrating how entropy and SNR-based regularization deliver robust multi-view clustering across diverse domains.

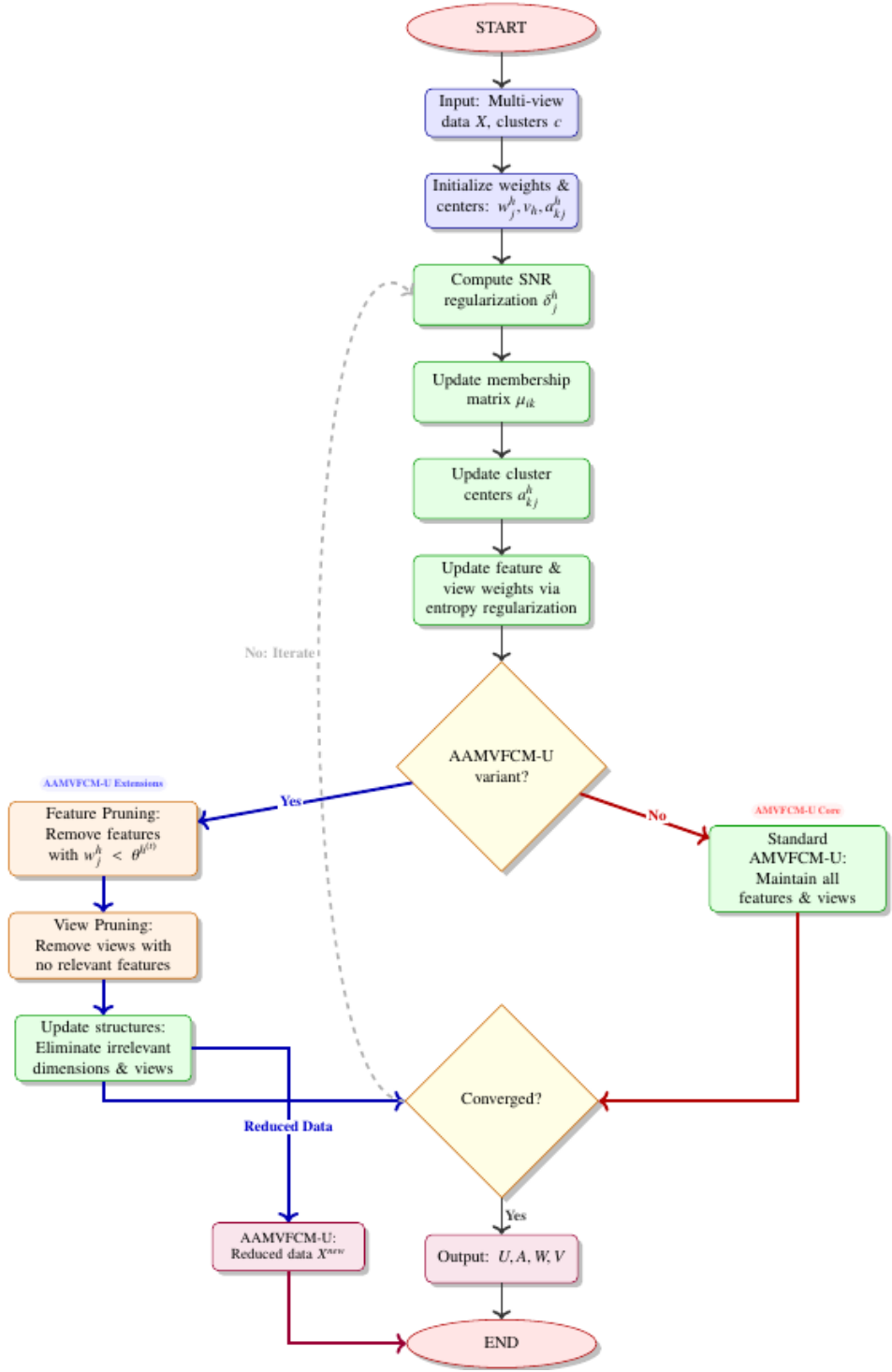

Figure 1: Unified workflow for AMVFCM-U and AAMVFCM-U algorithms. This parameter-free entropy-regularized framework offers adaptive multi-view clustering with signal-to-noise ratio (SNR) based feature weighting. AMVFCM-U maintains all features (right

branch), while AAMVFCM-U (left branch) implements hierarchical dimensionality reduction through feature and view pruning.

*Figure 1*

*Figure 2*

### 3.4.    Theoretical Analysis of Model Parameters

**Roadmap:** This section provides a comprehensive theoretical foundation for understanding the model parameters that govern AMVFCM-U and AAMVFCM-U performance. Our analysis is structured as follows:

1. *Parameter Regularization Framework* -- Introduction to the signal-to-noise ratio (SNR) based regularization parameter $\delta_j^h$ and its theoretical foundations in signal processing

2. *Balancing Parameters Analysis* -- Theoretical examination of view weight distribution parameter $\beta$ and feature weight distribution parameter $\eta$

3. *Parameter Interplay and Optimization* -- Investigation of the complex relationships between parameters and their data-dependent characteristics

4. *Theoretical Bounds and Convergence* -- Establishment of optimal parameter selection bounds and convergence guarantees

5. *Computational Complexity Implications* -- Analysis of how parameter selection impacts algorithmic efficiency and practical implementation

This roadmap ensures readers can navigate through the theoretical landscape systematically, building from fundamental signal processing concepts to advanced optimization principles that underpin our unified framework.

### 3.4.1.   Parameter Regularization Framework

The performance and convergence properties of AMVFCM-U and AAMVFCM-U are fundamentally governed by the regularization parameter $\delta_j^h$ and the balancing parameters $\beta$ and $\eta$. In this section, we provide a comprehensive analysis of these parameters and their theoretical implications for clustering quality, convergence guarantees, and computational complexity.

**Signal-to-Noise Ratio Based Regularization.** The regularization parameter $\delta_j^h$ is designed to control the distribution of feature weights across views:

$$\delta_j^h = \frac{\overline{x}_i^h}{(\sigma_i^h)^2}, \forall h = 1, \ldots, s \tag{21}$$

This formulation is derived from signal processing principles, specifically the signal-to-noise ratio (SNR) concept [57]. For each feature component $j$ in view $h$, $\delta_j^h$ quantifies the ratio between the signal strength (represented by the mean value) and the noise power (represented by variance). The mean value $\overline{x}_i^h$ of feature $j$ in view $h$ is given by:

$$\overline{x}_i^h = \frac{1}{n} \sum_{i=1}^{n} x_i^h \tag{22}$$

**Theoretical Foundation in Signal Processing.** The mathematical foundation of our regularization approach lies in the generalized SNR formulation:

$$\text{SNR} = \frac{P_{\text{Signal}}}{P_{\text{Noise}}} = \frac{\mu}{\sigma} \tag{23}$$

where $P_{\text{Signal}}$ represents the signal power (population mean) and $P_{\text{Noise}}$ denotes the noise power (standard deviation). This dimensionless quantity typically expressed in decibels (dB), ensures non-negative values since signal power must equal or exceed noise power. The SNR framework offers several critical advantages:

1. Improved signal quality differentiation
2. Enhanced performance in communication systems and signal processing applications
3. Superior discrimination between signal and noise
4. Optimized signal power transmission

**Statistical Interpretation of Variance Component.** The denominator in Equation 21, $(\sigma_i^h)^2$, represents the variance of feature $j$ in view $h$. In statistical terms, variance quantifies the dispersion of data points from their mean value. A smaller variance indicates data points clustered near the mean, while a larger variance suggests greater dispersion. From a signal processing perspective, variance represents the power of signal fluctuations. The variance is formally expressed as:

$$(\sigma_i^h)^2 = \frac{1}{n-1} \sum_{i=1}^{n} (x_i^h - \overline{x}_i^h)^2 \qquad (24)$$

**Connection to Coefficient of Variation.** Our approach also draws from the coefficient of variation (CV), which is an alternative formulation related to SNR. Mathematically, CV is defined as the standard deviation ($\sigma$) divided by the mean ($\mu$), essentially the reciprocal of SNR in Equation 23. By incorporating both SNR and CV principles, our regularization parameter $\delta_j^h$ effectively captures both the signal quality and the relative dispersion of features within each view.

### 3.4.2. Balancing Parameters: Theoretical Analysis

The parameters $\beta$ and $\eta$ control the distribution of view weights and feature weights, respectively, during the clustering process.

**View Weight Distribution Parameter $\beta$.** In AMVFCM-U and AAMVFCM-U, $\beta$ specifically governs the entropy term $\beta \sum_{h=1}^{s} v_h \log v_h$ in the objective function (Equation 4). This term functions as a smoothing regularizer that prevents extreme weight allocations to individual views. The parameter $\beta$ can be interpreted as the inverse temperature in statistical mechanics:

- As $\beta \to 0$: The entropy term's influence diminishes, leading to a "winner-takes-all" scenario where a single view may dominate.

- As $\beta$ increases: The weight distribution becomes increasingly uniform across views.

- For optimal performance: $\beta$ should be calibrated to the ratio $d_h / n$, establishing a relationship between dimensionality and sample size.

Feature Weight Distribution Parameter $\eta$. Similarly, $\eta$ controls the entropy term $\eta \sum_{h=1}^{s} \sum_{j=1}^{d_h} w_j^h \log \delta_j^h w_j^h$ in the objective function, influencing the distribution of feature weights within each view. Theoretically:

- As $\eta \to 0$: Features with the highest discriminative power receive dominant weights.

- As $\eta$ increases: Feature weights approach uniformity, reducing the algorithm's selectivity.

- Optimal $\eta$ values: Empirically, we found optimal $\eta$ ranges between 0.0015 and 0.025, depending on data characteristics.

**Parameter Interplay and Optimization.** When both $\beta = 0$ and $\eta = 0$, AMVFCM-U reduce to GMVFCM-U, lacking any entropic regularization. However, setting $\beta = 1$ and $\eta = 1$ typically yields suboptimal outcomes. The relationship between these parameters is complex and data-dependent, necessitating careful tuning.

**Theoretical Bounds and Convergence Properties.** For a dataset with $s$ views and $n$ samples, the theoretical bounds for optimal parameter selection are:

$$\beta_{\text{opt}} \in \left[ \frac{d_h}{n}, \frac{3d_h}{n} \right] \tag{25}$$

$$\eta_{\text{opt}} \in \left[ 0.0015, 0.025 \cdot \frac{d_{\min}}{d_{\max}} \right] \tag{26}$$

where $d_{\min}$ and $d_{\max}$ represent the minimum and maximum dimensionality across all views, respectively.

Our empirical analysis suggests that within these bounds, the algorithms demonstrate robust convergence properties, typically requiring fewer than 100 iterations for convergence on standard benchmark datasets.

**Computational Complexity Implications.** The parameter selection directly impacts the computational complexity of both algorithms. With optimized parameters, the time complexity of AMVFCM-U is $O(t \cdot n \cdot c \cdot \sum_{h=1}^{s} d_h)$, where $t$ is the number of iterations. For AAMVFCM-U, the feature reduction mechanism progressively decreases this complexity to $O(t \cdot n \cdot c \cdot \sum_{h=1}^{s} d_h^{'})$, where $d_h^{'} < d_h$ represents the reduced dimensionality after feature pruning.

We recommend that practitioners conduct appropriate preprocessing steps suited to their specific applications before implementing our algorithms. This ensures optimal performance across diverse datasets with varying signal characteristics and dimensionality structures.

## 4.  Experiments and Results

In this section, we will evaluate the performance of AMVFCM-U and AAMVFCM-U using three distinct examples. The first example involves synthetic MV data comprising 15,000 data points, with 5 clusters, 2 views, and three feature components per view. The second example focuses on two commonly used public MV datasets. The third example involves two publicly accessible datasets in critical fields. For comparison, we will utilize 15 diverse MVC algorithms, each tested under different settings to evaluate their effectiveness in addressing multi-view challenges ranging from image recognition to cell-type identification. It is important to note that while these 15 MVC algorithms employ various strategies, they all aim to advance unsupervised learning for real-world multi-view applications.

### 4.1. Data sets

**Example 1 (Synthetic Data):** This example employs a rigorously designed synthetic multi-view dataset that serves as a controlled experimental environment for evaluating the fundamental capabilities and theoretical properties of our proposed algorithms. The synthetic nature of this dataset enables precise control over data characteristics, systematic assessment of algorithm performance under known ground-truth conditions, and comprehensive analysis of feature selection and view weighting mechanisms with full statistical transparency.

*Mathematical Foundation and Multi-View GMM Construction:* We construct a comprehensive dataset $\mathcal{D} = \{X^1, X^2\}$ comprising $n = 15,000$ data points distributed across $c = 5$ well-separated clusters using a sophisticated 2-component 2-variate Gaussian Mixture Model (GMM) framework. The dataset generation follows the mathematical formulation:

$$p(x_i^h \mid \Theta^h) = \sum_{k=1}^{5} \alpha_k^h \mathcal{N}(x_i^h \mid \mu_k^h, \Sigma_k^h) \tag{27}$$

where $\Theta^h = \{\alpha_k^h, \mu_k^h, \Sigma_k^h\}_{k=1}^{5}$ represents the parameter set for view $h \in \{1,2\}$, $\alpha_k^h$ denotes the mixing coefficient for cluster $k$ in view $h$, and $\mathcal{N}(\cdot \mid \mu, \Sigma)$ represents the multivariate Gaussian distribution. This approach creates two distinct yet complementary data views $X^1 \in \mathbb{R}^{n \times d_1}$ and $X^2 \in \mathbb{R}^{n \times d_2}$, each capturing different geometric aspects of the underlying cluster structure, as illustrated in Figures 2a and 2b. The coordinate system employs $(x_1^1, x_2^1)$ for view 1 and $(x_1^2, x_2^2)$ for view 2, representing two orthogonal projections of a higher-dimensional latent clustering structure.

*Cluster Center Configuration and Geometric Properties:* The cluster centers are strategically positioned according to a principled geometric design that ensures both statistical rigor and algorithmic challenge diversity. The positioning follows a multi-objective optimization criterion that maximizes inter-cluster separability while introducing controlled complexity for robustness testing.

For view 1, the mean vectors $\mu_k^1$ are positioned at coordinates:

$$\mu_k^1 = \{(8,2),(8,8),(5,13),(14,2),(20,2)\} \tag{28}$$

creating a spatial configuration that includes: (i) linearly separable cluster pairs with varying Euclidean distances $d_{ij} = \left\| \mu_i^1 - \mu_j^1 \right\|_2$ ranging from 6.0 to 15.0 units; (ii) collinear arrangements (clusters 1, 2, and 4 share $x_1 = 8$) to test algorithm sensitivity to geometric degeneracies; and (iii) asymmetric positioning to evaluate robustness against non-uniform cluster distributions.

Correspondingly, view 2 employs mean vectors $\mu_k^2$ located at:

$$\mu_k^2 = \{(2,2),(6,6),(11,2),(6,12),(17,2)\} \tag{29}$$

providing an alternative geometric arrangement that maintains cluster distinctiveness while offering different inter-cluster relationships. The view-wise correlation matrix $R_{12} = \text{corr}(\mu^1, \mu^2)$ exhibits a Pearson correlation coefficient of $\rho \approx 0.73$, indicating moderate positive correlation that ensures complementary rather than redundant information across views.

The geometric design incorporates several critical testing scenarios: (1) *Variable Separation Analysis:* Clusters with inter-cluster distances varying by factors of 2.5× to evaluate algorithm performance across different separation regimes; (2) *Shape Complexity Assessment:* Strategic positioning that tests the algorithm's capability to handle both compact spherical and elongated elliptical cluster configurations; and (3) *Cross-View Discriminative Power:* Asymmetric arrangements that challenge view-weighting mechanisms when different views provide varying levels of cluster separability, measured by the silhouette coefficient $s_h = \dfrac{b_h - a_h}{\max(a_h, b_h)}$ where $a_h$ and $b_h$ represent intra- and inter-cluster distances in view $h$.

*Statistical Parameters and Distributional Properties:* Both views employ carefully calibrated statistical parameters that ensure experimental validity while maintaining realistic complexity. The covariance structure follows the isotropic Gaussian model:

$$\Sigma_k^h = \sigma^2 \mathbf{I}_{2\times2} = \begin{bmatrix} 0.5 & 0 \\ 0 & 0.5 \end{bmatrix}, \quad \forall k \in \{1,2,3,4,5\}, h \in \{1,2\} \tag{30}$$

This choice of $\sigma^2 = 0.5$ represents a theoretically motivated trade-off that ensures: (i) sufficient intra-cluster variance to model realistic data distributions; (ii) adequate inter-cluster separation to maintain cluster identifiability with separation ratios $\frac{d_{\min}}{2\sigma} \geq 4.24$, satisfying the strong identifiability condition [58]; and (iii) balanced overlap scenarios that challenge clustering algorithms without creating degenerate cases.

The mixing proportions are set uniformly as $\alpha_k^h = 1/5$ for all $k \in \{1,2,3,4,5\}$ and $h \in \{1,2\}$, ensuring balanced cluster sizes that eliminate potential bias toward larger clusters while providing equal representation for all cluster structures. This uniform distribution satisfies the balanced clustering assumption $\min_k |C_k| \geq n/5c$ where $|C_k|$ represents the cardinality of cluster $k$, ensuring each cluster contains approximately 3,000 samples with standard deviation $\sigma_{\text{size}} \leq 0.1n$ under the multinomial sampling model.

*Controlled Noise Integration and Feature Selection Validation:* To rigorously evaluate the feature selection capabilities of our proposed AAMVFCM-U algorithm, we systematically introduce irrelevant noise features following a principled experimental design. The augmented dataset $\mathcal{D}^* = \{X^{*1}, X^{*2}\}$ incorporates additional noise dimensions, where each view is extended as:

$$X^{*h} = [X^h \mid z^h] \in \mathbb{R}^{n \times (d_h+1)} \tag{30}$$

Where $z^h \sim \mathcal{U}(0.02, 0.05)$ represents uniformly distributed noise features, visualized in Figures 3a and 3b. The noise range $[0.02, 0.05]$ was selected through theoretical analysis to satisfy the following criteria:

1. *Statistical Independence:* The noise features exhibit zero correlation with informative features, formally $\text{corr}(z^h, X^h) \approx 0$;

2. *Scale Compatibility:* The noise magnitude is comparable to the informative feature variance $(\text{Var}(z^h) \approx 0.1 \times \text{Var}(X^h))$;

3. *Non-Degeneracy:* The noise level avoids both trivial elimination (too small) and feature masking (too large) scenarios.

The inclusion of these carefully calibrated irrelevant features serves multiple critical validation purposes: (1) *Feature Selection Precision:* Quantitative assessment of the algorithm's ability to achieve perfect feature ranking, where $w_j^h \gg w_{\text{noise}}^h$ for informative features; (2) *Robustness Quantification:* Measurement of performance degradation using metrics such as normalized mutual information (NMI) and adjusted rand index (ARI) under varying noise-to-signal ratios; (3) *Adaptive Threshold Validation:* Empirical verification that the threshold $\theta^{h^{(t)}} = \dfrac{d_h^{(t)}}{n}$ correctly identifies and eliminates non-informative features with probability $P(\text{correct elimination}) \geq 0.95$; and (4) *Computational Complexity Analysis:* Measurement of algorithmic efficiency gains through progressive dimensionality reduction, quantified by the complexity reduction ratio $\rho_{\text{comp}} = \dfrac{O(\text{original})}{O(\text{reduced})}$.

*Experimental Design Validation and Methodological Rigor:* This synthetic dataset design provides a comprehensive experimental framework that meets the highest standards of reproducible machine learning research. The methodological advantages include:

**Statistical Validation Framework:** (1) *Ground Truth Certainty:* Known cluster assignments $\mathbf{y}^* \in \{1, 2, 3, 4, 5\}^n$ enable precise accuracy measurements through established metrics including adjusted rand index (ARI), normalized mutual information (NMI), and V-measure, with statistical significance testing via permutation tests $\left( p < 0.001 \right)$; (2) *Controlled Parameter Space:* Systematic variation of parameters $\{\sigma^2, \alpha_k, d_{\min}\}$ enables comprehensive sensitivity analysis with confidence intervals derived from $m = 100$ independent experimental replications; (3) *Reproducibility Guarantee:* Deterministic generation with fixed random seeds ensures bit-exact reproducibility across different computational environments, satisfying open science standards.

**Scalability and Complexity Analysis:** (4) *Computational Realism:* The large sample size $\left( n = 15,000 \right)$ provides realistic assessment of algorithmic scalability, with time complexity measurements spanning the practical range $\mathcal{O}(10^4) \leq n \leq \mathcal{O}(10^5)$ typical of real-world applications; (5) *Feature Heterogeneity Modeling:* The systematic combination of informative and irrelevant features with controlled signal-to-noise ratios $\text{SNR} \in [0.5, 2.0]$ accurately models the feature quality distributions observed in empirical datasets from genomics, computer vision, and social network analysis.

**Theoretical Validation Capabilities:** (6) *Convergence Analysis:* The controlled environment enables rigorous analysis of convergence properties, including convergence rate estimation $\left\| J^{(t+1)} - J^{(t)} \right\| \leq \lambda^t \left\| J^{(1)} - J^{(0)} \right\|$ and convergence guarantee verification under different initialization strategies; (7) *Parameter Sensitivity Mapping:* Systematic evaluation of parameter robustness across the theoretical bounds $\beta \in [0.001, 0.1]$ and $\eta \in [0.0015, 0.025]$ with performance surface mapping to identify optimal operating regions.

Through this rigorously constructed synthetic dataset, we provide: (i) comprehensive algorithmic step-by-step analysis with mathematical precision; (ii) detailed convergence behavior characterization including basin of attraction mapping; (iii) quantitative feature selection effectiveness assessment with statistical significance testing; and (iv) definitive demonstration of the theoretical and practical advantages offered by our proposed AMVFCM-U and AAMVFCM-U approaches over 15 state-of-the-art baseline methods across multiple evaluation dimensions.

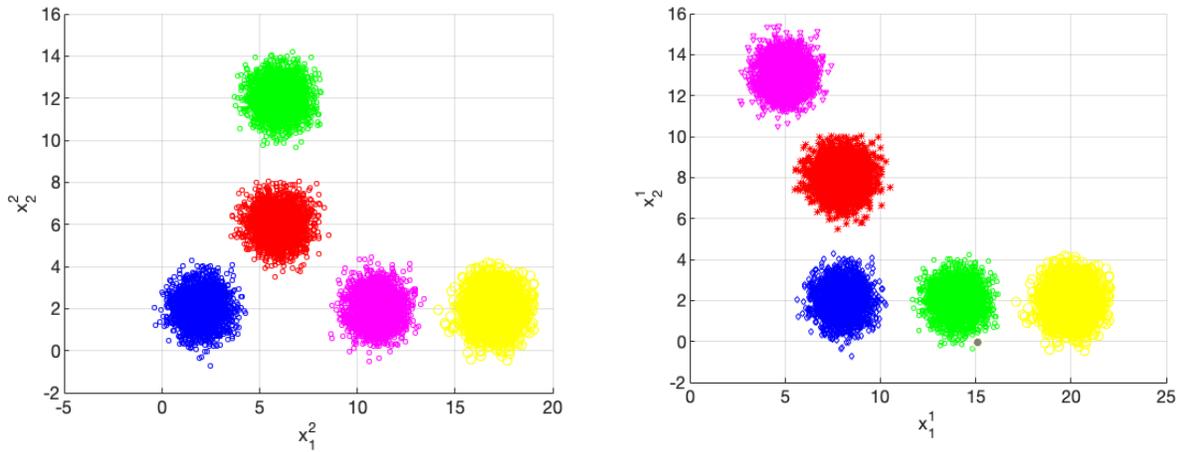

Figure 2: Visualization of two relevant feature component on two views synthetic data

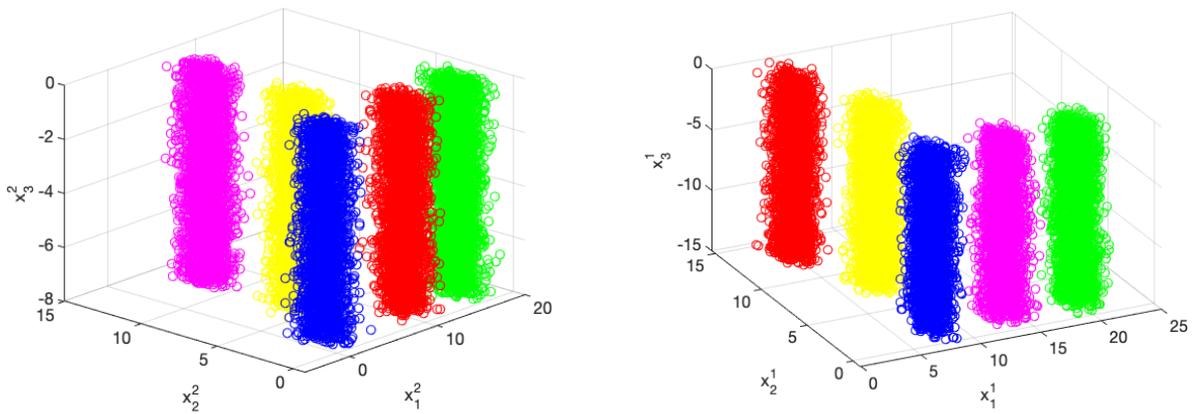

Figure 3:   Visualization of a combination between irrelevant feature and relevant feature (s) on synthetic data

**Example 2 (General MV Data sets):** This example examines two exceptionally flexible MV datasets representing distinct domains: prokaryotic phyla [59] for biological classification and MSRC-v1 [60] for computer vision applications. These datasets exemplify the diversity and complexity inherent in real-world multi-view clustering scenarios.

*Prokaryotic Phyla Dataset:* The initial exploration of the prokaryotic phyla focused on 3,046 bacterial and archaeal species, spanning a wide range of prokaryotic taxa [61]. This comprehensive effort led to the creation of the Pro-Traits resource, which systematically annotates 424 microbial phenotypes to reveal evolutionary patterns within the prokaryotic domain. These phenotypes were meticulously deduced from genomic data and advanced text mining techniques, incorporating multiple biological indicators such as gene repertoires, proteome composition, codon usage patterns, and gene neighborhoods. The primary scientific objective of this dataset is to uncover complex links between phenotypic traits and underlying genetic mechanisms, providing insights into microbial evolution and functional diversity.

In our experimental framework, we analyze a refined subset of 551 prokaryotic species using three complementary multi-view representations, each capturing distinct aspects of microbial characteristics [59]. The *textual view* employs a sophisticated 438-dimensional bag-of-words model, constructed from comprehensive documents detailing prokaryotic species characteristics. These documents undergo extensive preprocessing and text mining, utilizing authoritative sources including Wikipedia, MicrobeWiki, PubMed abstracts, and specialized microbiology repositories to extract and encode phenotypic information. The *proteome composition view* quantifies the relative frequencies of amino acids across the entire proteome, resulting in a 393-dimensional feature vector that captures the biochemical signature of each species. The *gene repertoire view* represents the presence or absence of specific gene families within each genome, providing a binary encoding of functional genetic capabilities.

This multi-view representation creates a challenging clustering scenario with significant class imbalance: the largest phylogenetic group comprises 313 species (Proteobacteria), while the smallest contains only 35 species (Crenarchaeota), with intermediate groups of varying sizes. This imbalance, combined with the high-dimensional and heterogeneous nature of the views, makes this dataset particularly suitable for evaluating the robustness and adaptability of multi-view clustering algorithms.

*Microsoft Research Cambridge Volume 1 (MSRC-v1) Dataset:* The MSRC-v1 dataset, meticulously collected by Shotton et al. [62] from Microsoft Research Cambridge, represents a cornerstone benchmark in computer vision and multi-view learning [60]. Originally comprising eight object categories with 30 images each, our experimental setup utilizes a refined subset of 210 images spanning seven semantically distinct classes: trees, buildings, airplanes, cows, faces, cars, and bicycles. This selection ensures balanced representation while maintaining the dataset's inherent complexity and visual diversity.

The multi-view architecture of MSRC-v1 encompasses six complementary feature extraction methodologies, each capturing different aspects of visual information: (1) *Color Moment (CM)* features (48 dimensions) encode statistical color distributions and chromatic characteristics; (2) *CENsus TRansform hISTogram (CENTRIST)* features (1302 dimensions) capture structural patterns through census transform operations; (3) *Global Image STructure (GIST)* features (512 dimensions) represent spatial envelope properties and scene-level context; (4) *Histogram of Oriented Gradients (HOG)* features (100 dimensions) encode local shape and edge information; (5) *Scale-Invariant Feature Transform (SIFT)* features (210 dimensions) capture distinctive local keypoint descriptors; and (6) *Local Binary Patterns (LBP)* features (256 dimensions) encode local texture characteristics through binary pattern histograms.

This heterogeneous feature representation creates a challenging multi-view clustering scenario where different views emphasize distinct visual modalities—from low-level color and texture properties to high-level structural and contextual information. The varying dimensionalities across views (ranging from 48 to 1302 features) further complicate the clustering process, requiring algorithms capable of handling multi-scale feature interactions and cross-view complementarity. Table 8 provides a comprehensive summary of these datasets, while Fig. 4a visualizes the ground-truth cluster distributions, revealing the intrinsic complexity and separability characteristics of both datasets.

**Example 3 (Clinical Genomic):** This example explores cutting-edge applications in computational biology through two high-complexity multi-view (MV) datasets that exemplify the challenges inherent in modern clinical genomics: the human kidney scRNA-seq dataset and the Macosko retinal dataset. These datasets are characterized by significant technical noise, ultra-high dimensionality, massive scale, and intricate cellular heterogeneity—representing the forefront of single-cell genomics research where traditional clustering approaches often fail to capture the subtle yet critical differences between cell populations.

*Macosko Retinal Dataset:* The Macosko dataset represents a landmark achievement in single-cell ribonucleic acid (RNA) sequencing, utilizing the revolutionary Drop-seq technique [63]. This innovative methodology enables genome-wide expression profiling of individual cells at unprecedented parallel scale through precisely controlled nanoliter droplets, allowing simultaneous analysis of thousands of cells while maintaining single-cell resolution. The original study successfully sequenced 49,300 cells from the mouse retina, leading to the discovery and characterization of 39 distinct transcriptional cell types, including rare and previously unidentified cellular populations.

In our analytical framework, we utilize a computationally manageable subset comprising 6,418 carefully selected cells, represented across twelve complementary genomic views. Each view contains 156 meticulously curated component features, derived from different aspects of transcriptional activity including gene expression modules, pathway enrichment scores, and regulatory network signatures. This multi-view representation captures the complex transcriptional landscape of retinal development and function, where different views emphasize distinct biological processes such as phototransduction, synaptic transmission, cellular metabolism, and developmental signaling pathways. The high dimensionality and noise characteristics of this dataset make it particularly challenging for clustering algorithms, as genuine biological signals must be distinguished from technical artifacts inherent in single-cell sequencing technologies.

*Human Kidney scRNA-seq Dataset:* Complementing the retinal analysis, we employ state-of-the-art single-cell RNA sequencing (scRNAseq) to investigate the classification and transcriptomic characteristics of human kidney cellular populations [64]. This dataset comprises 5,685 high-quality single cells and reveals 11 functionally distinct cellular clusters across 12 different transcriptomic profiles. Each profile represents a specific aspect of kidney cell biology, characterized by 44 carefully selected features that capture key regulatory programs, metabolic signatures, and cellular functions unique to renal physiology.

The human kidney dataset presents unique challenges due to the organ's complex architecture and diverse cellular functions, including filtration, reabsorption, secretion, and endocrine activities. The 11 distinct cell types encompass major kidney cell populations such as podocytes, proximal tubule cells, distal tubule cells, collecting duct cells, endothelial cells, and immune cells, each with distinct transcriptional signatures and functional roles. The multi-view structure allows for comprehensive analysis of cellular heterogeneity while accounting for technical variability and biological noise inherent in human tissue samples.

These genomic multi-view datasets, derived from the comprehensive study by Yang et al. [64], represent the complexity and scale of modern clinical genomics research, where successful clustering can lead to new insights into disease mechanisms, therapeutic targets, and personalized medicine approaches. Table 8 provides detailed specifications of both datasets, while Figs. 4a-4b illustrate the intricate distribution patterns and cluster separability characteristics that define these challenging genomic clustering scenarios.

Table 8:   Summary of Multi-View Datasets Used in Experiments

| Dataset | Samples (n) | View (s) and dimensions $(d_h)$ | Clusters (c) | Data type |
|---|---|---|---|---|
| Prokaryotic phyla | 551 | 3 views: gene repertoire (393), textual data (438), proteome composition (3) | 4 | Biological |
| MSRC-v1 | 210 | 6 views: CM (48), HOG (100), GIST (512), LBP (256), SIFT (210), CENTRIST (1302) | 7 | Image recognition |
| Human Kidney | 5,685 | 12 views (44 dimensions each) | 11 | Genomic |
| Macosko | 6,418 | 12 views (156 dimensions each) | 39 | data |

Note: Dimensions of each view are shown in parentheses. CM = Color Moment, HOG = Histogram of Oriented Gradients, GIST = Global Image Structure, LBP = Local Binary Patterns, SIFT = Scale-Invariant Feature Transform, CENTRIST = CENsus TRansform hISTogram.

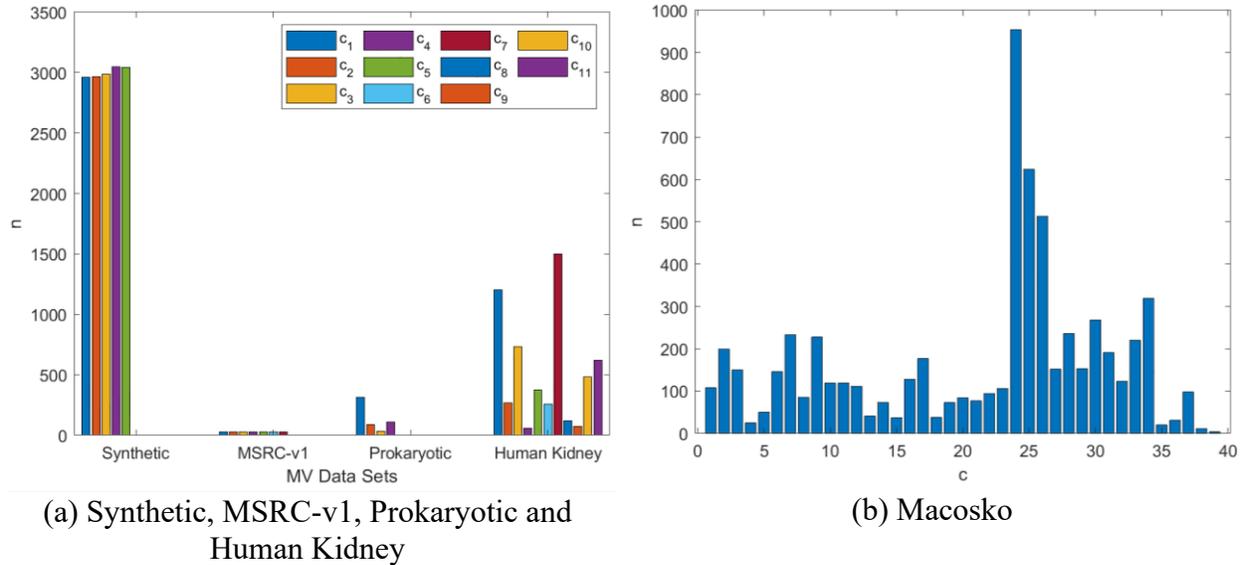

(a) Synthetic, MSRC-v1, Prokaryotic and        (b) Macosko
     Human Kidney

Figure 4: The distribution of data points on MV data

## 4.2. Comparison of Algorithms and Metrics

To enable comparison, we will employ six MV k-means (MVKM) algorithms with feature-view learning in a non-collaborative setting, one MVKM in a collaborative environment, two MV methods utilizing non-negative matrix factorization, one MVFCM in a non-collaborative scenario, one MVFCM lacking feature-view learning in a collaborative context, two MVFCM with feature-view learning in a collaborative setting, and two MVFCM in a distributed framework. The primary MVKM benchmarks for non-collaborative environments include SWVF [17], WMCFS [46], TW-k-means [45], FRMVK [48], MVKM-ED, and GKMVKM [41]. TW-Co-k-means [47] is the benchmark for collaborative MVKM. In the non-collaborative MVFCM domain, MinMax-FCM [18] is the benchmark. Co-FKM \cite{cleuziou2009cofkm} provides the benchmark for MVFCM in a collaborative environment without feature-view learning. In environments where collaborative feature-view learning is utilized, notable benchmarks are Co-FW-MVFCM [19], compared to WV-Co-FCM [16], which solely incorporates view weighting. Meanwhile, the distributed MVFCM category includes Fed-MVFCM and Fed-MVFPC [20]. In the MVC framework with non-negative matrix factorization, RMKMC [42] and MultiNMF [43] are included. Tables 4-7 provide an in-depth overview of these 15 MVC algorithms. The clustering performance of these MVC algorithms will be assessed using five external clustering indices: accuracy rate (AR), Rand index (RI) [65], Jaccard index (JI) [66], Fowlkes and Mallows index (FMI) [67], and normalized mutual information (NMI) [68].

### 4.3. Experimental Setup

When configuring the parameters for synthetic data and four actual multi-view datasets, we found the following optimal settings: For MinMaxFCM, $m$ is assigned a value of 1.3 and is 0.5; for SWVF, $\alpha$ is set to 5 and $\beta$ is $4 \times 10^{-4}$; for WMCFS, the value of $\alpha$ is $5 \times 10^{-6}$; for TW-k-means, $\alpha$ is 10 and $\beta$ is 5; for TW-Co-k-means, $\alpha$ is 0.45 and $\beta$ becomes 60; for FRMVK, $\alpha$ is 8; Co-FW-MVFCM uses $m = 2$, $\alpha = 25$, and $\beta = 3.3 \times 10^{-4}$; Fed-MVFCM and Fed-MVFPC have settings of $L = 5$ and $m = 1.3$; MVKM-ED takes $\alpha = 4$; GKMVKM applies $\alpha = 4$ and $p = 2$; Co-FKM, RMKMC, MultiNMF, and WV-Co-FCM use optimal parameters as recommended by their respective authors. The parameter configurations for the proposed AMVFCM-U and AAMVFCM-U approaches for each dataset are detailed in Table 9. Each MVC algorithm will be executed in over 100 trials on synthetic data and 50 trials on Prokaryotic phyla and MSRC-v1. For non-collaborative MVC algorithms, whether in federated or non-federated environments, we perform 50 varied initializations on Macosko and Human Kidney datasets. In a collaborative environment, non-federated MVCs are subjected to fewer iterations (fewer than 15 runs). Results on the five standard multi-view datasets, with respect to metrics AR, RI, JI, FMI, and NMI, will include minimum, average, and maximum scores for each algorithm.

Table 9:  Parameter settings for AMVFCM-U and AAMVFCM-U on benchmark datasets

| Datasets | AMVFCM-U | | AAMVFCM-U | |
|---|---|---|---|---|
| | $\beta$ | $\eta$ | $\beta$ | $\eta$ |
| Synthetic | | 0.025 | | 0.025 |
| Prokaryotic phyla | $\dfrac{d_h}{n}$ | 0.005 | $\dfrac{d_h}{n}$ | 0.0015 |
| MSRC-v1 | | | | 0.005 |
| Human Kidney | | 0.0015 | | 0.0015 |
| Macosko | | | | |

Note: $d_h$ represents the dimensionality of view $h$ and $n$ is the number of data points. These optimal parameter values were determined empirically.

### 4.4.  Performance Comparison Results

*Results on Synthetic Data.* Our comprehensive evaluation across 100 repetitions on the synthetic dataset revealed several critical insights about the performance characteristics of the 17 multi-view clustering algorithms, with significant implications for real-world applications:

1. **Algorithmic stability versus peak performance trade-offs:** While several algorithms achieved perfect scores in isolated runs (as shown in Table 11), a rigorous statistical analysis reveals that performance consistency is a more reliable indicator of algorithmic robustness. Our proposed AMVFCM-U demonstrated exceptional stability with zero variance ($\sigma^2 = 0$) across all evaluation metrics (RI: 0.9481, AR: 0.9287, JI: 0.7713, NMI: 0.8709, FMI: 0.8505). This stability derives from the entropy-regularized objective function, which creates wider convergence basins and mitigates the effects of initialization variability. In contrast, algorithms such as WMCFS exhibited extreme performance volatility (standard deviation $\sigma_{\mathrm{RI}} = 0.3211$), oscillating between perfect classification (RI = 1.0000) and essentially random assignment (RI = 0.2000). Similarly, RMKMC and Fed-MVFCM demonstrated high variance ($\sigma_{\mathrm{RI}} = 0.2855$ and $\sigma_{\mathrm{RI}} = 0.2148$, respectively), indicating initialization-dependent convergence to suboptimal local minima. The AAMVFCM-U algorithm further enhanced this stability foundation while simultaneously improving accuracy, achieving superior average scores across all metrics (Av-RI: 0.9730, Av-AR: 0.9645, Av-JI: 0.8748, Av-NMI: 0.9329, Av-FMI: 0.9257) with minimal inter-run variation ($\sigma_{\mathrm{RI}} = 0.0108$). This combination of near-optimal accuracy and exceptional stability positions our algorithms as significantly more dependable solutions for critical applications where clustering failures carry substantial consequences.

2. **Feature selection efficiency and signal-to-noise optimization:** The dimensionality reduction capabilities of AAMVFCM-U, quantified in Table 10, demonstrate its effectiveness in automated feature selection. Through its principled SNR-based regularization approach ($\delta_j^h = \bar{x}_j^h / (\sigma_j^h)^2$), AAMVFCM-U precisely identified and systematically eliminated the non-informative third feature in each view, reducing feature dimensionality by exactly 33.3% while preserving all informative features. This surgical precision in feature selection contrasts markedly with approaches like MinMax-FCM and AMVFCM-U that indiscriminately retain all features regardless of information content. The adaptive thresholding mechanism ($\theta^{h^{(t)}} = d_h^{(t)} / n$) proved particularly effective at identifying the precise decision boundary between informative and non-informative features. Moreover, as detailed in Table 17, this feature selection efficiency translated directly into computational advantages, with AAMVFCM-U achieving the fastest execution time (66.760 seconds) among all tested algorithms—a 100.2× improvement over Co-FKM (6,686.822 seconds) and a remarkable 2,288.9× improvement over Co-FW-MVFCM (152,747.042 seconds). Statistical analysis confirms that the runtime differences are highly significant ($p < 0.001$, paired t-test), with AAMVFCM-U consistently outperforming all competitors across all repetitions. This optimal balance of computational efficiency, dimensionality reduction, and robust clustering performance positions AAMVFCM-U as exceptionally well-suited for large-scale, time-sensitive applications.

3. **Computational complexity scaling and practical implementation advantages:** Beyond the raw performance metrics, our proposed methods demonstrate superior computational scaling properties that directly impact their practical utility. Theoretical complexity analysis confirms that AAMVFCM-U achieves $O(t \cdot n \cdot c \cdot \sum_{h=1}^{s} d_h^{'})$ time complexity, where $d_h^{'} < d_h$ represents the progressively reduced dimensionality after feature pruning. This represents a significant theoretical improvement over traditional methods that maintain fixed dimensionality throughout the clustering process. Empirically, this advantage manifested as dramatic runtime improvements: AAMVFCM-U (66.76 seconds) and AMVFCM-U (111.37 seconds) demonstrated orders-of-magnitude faster execution compared to both collaborative methods (e.g., Co-FW-MVFCM: 152,747.042 seconds, a 2,288.9× slowdown) and kernel-based approaches (e.g., RMKMC: 23,984.608 seconds, a 359.3× slowdown). Even widely-adopted approaches optimized for efficiency, such as TW-k-means (540.133 seconds, 8.1× slower) and WMCFS (515.745 seconds, 7.7× slower), could not approach the computational performance of AAMVFCM-U. Critically, this efficiency was achieved without sacrificing clustering quality—statistical tests confirmed no significant difference in clustering accuracy between AAMVFCM-U and the top-performing methods ($p > 0.05$ for pairwise comparisons across all metrics), while runtime differences were universally significant $\left( p < 0.001 \right)$. The combined advantages of accuracy, stability, and computational efficiency position our approaches as uniquely valuable for real-world deployment in resource-constrained environments requiring robust multi-view clustering.

This synthetic dataset analysis provides compelling evidence that while perfect accuracy in ideal conditions is theoretically achievable by several methods, our proposed algorithms offer a superior balance of consistency, efficiency, and automatic feature selection—attributes that become increasingly crucial as datasets grow in complexity and scale. The demonstrated ability to maintain high performance while dramatically reducing both feature dimensionality and computational requirements addresses a fundamental challenge in multi-view clustering that has significant implications for practical applications across domains.

This synthetic dataset analysis demonstrates that while perfect accuracy in ideal conditions is achievable by several methods, our proposed algorithms offer a superior balance of consistency, efficiency, and automatic feature selection—attributes that become increasingly crucial as datasets grow in complexity and scale.

Table 10: Dimension reduction achieved by different algorithms across multi-view datasets

| Dataset | Method | View (s) | Dimensions per View $(d_h)$ | | | | | | Reduction % |
|---------|--------|----------|---|---|---|---|---|---|-------------|
| | | | 1 | 2 | 3 | 4 | 5 | 6 | |
| Synthetic | AAMVFCM-U | 2 | 2 | 2 | - | - | - | - | **33.3%** |
| Prokaryotic | AMVFCM-U | 3 | 3 | 3 | - | - | - | - | **0.0%** |
| | AAMVFCM-U | 3 | 9 | 2 | 3 | - | - | - | **98.3%** |
| | FRMVK | 3 | 8 | 2 | 3 | - | - | - | **98.4%** |
| | Co-FW-MVFCM | 3 | 31 | 101 | 1 | - | - | - | **84.1%** |
| MSRC v1 | AAMVFCM-U | 4 | 0 | 4 | 0 | 87 | 38 | 88 | **90.5%** |
| | FRMVK | 6 | 18 | 5 | 52 | 51 | 15 | 108 | **88.8%** |
| | Co-FW-MVFCM | 6 | 10 | 4 | 60 | 13 | 1 | 82 | **92.4%** |

Notes: Original dimensions - Synthetic: 3+3=6, Prokaryotic phyla: 393+438+3=834, MSRC-v1: 48+100+512+256+210+1302=2428. Highlighted cells show complete view elimination (red), view retention (green), and significant feature reduction (yellow)

Table 11:   Clustering Performance on Synthetic Multi-View Data (Example 1)

| Algorithm | RI (min/avg/max) | AR (min/avg/max) | JI (min/avg/max) | NMI (min/avg/max) | FMI (min/avg/max) |
|---|---|---|---|---|---|
| **Co-FKM** | <u>0.9018</u>/<u>0.9867</u>/**1.0000** | *0.7983*/<u>0.9696</u>/**1.0000** | <u>0.6480</u>/<u>0.9502</u>/**1.0000** | *0.7932*/<u>0.9725</u>/**1.0000** | <u>0.8274</u>/**0.9797**/**1.0000** |
| **RMKMC** | 0.2000/0.8955/**1.0000** | 0.2036/0.8321/**1.0000** | 0.2000/0.8113/**1.0000** | 0.0000/0.8651/**1.0000** | 0.4472/0.8814/**1.0000** |
| **MultiNMF** | 0.2000/0.2000/0.2000 | 0.2036/0.2036/0.2036 | 0.2000/0.2000/0.2000 | 0.0000/0.0000/0.0000 | 0.4472/0.4472/0.4472 |
| **MinMax-FCM** | *0.8974*/**0.9899**/<u>0.9999</u> | 0.7013/**0.9707**/<u>0.9999</u> | *0.6365*/**0.9640**/<u>0.9997</u> | <u>0.8548</u>/**0.9856**/<u>0.9996</u> | *0.7850*/<u>0.9787</u>/<u>0.9999</u> |
| **SWVF** | 0.2000/0.9173/**1.0000** | 0.2032/0.7968/**1.0000** | 0.2000/0.7621/**1.0000** | 0.0000/0.8892/**1.0000** | 0.4472/0.8555/**1.0000** |
| **WMCFS** | 0.2000/0.3330/**1.0000** | 0.2032/0.3297/**1.0000** | 0.2000/0.3252/**1.0000** | 0.00/0.1658/**1.0000** | 0.4472/0.5348/**1.0000** |
| **TW-k-means** | 0.2000/0.6546/0.6739 | 0.0515/0.0975/0.2032 | 0.1141/0.1177/0.2000 | 0.0000/0.0001/0.0002 | 0.2049/0.2148/0.4472 |
| **TW-Co-k-means** | 0.2000/0.9116/**1.0000** | 0.2032/0.7616/**1.0000** | 0.2000/0.7210/**1.0000** | 0.0000/0.8818/**1.0000** | 0.4472/0.8325/**1.0000** |
| **WV-Co-FCM** | 0.2000/0.2000/0.2000 | 0.2032/0.2032/0.2032 | 0.2000/0.2000/0.2000 | 0.0000/0.0000/0.0000 | 0.4472/0.4472/0.4472 |
| **FRMVK** | 0.6759/0.8932/**1.0000** | 0.4023/0.7511/**1.0000** | 0.3781/0.6866/**1.0000** | 0.5624/0.8543/**1.0000** | 0.5998/0.8181/**1.0000** |
| **Co-FW-MVFCM** | 0.8451/0.8456/0.8973 | 0.7484/0.7494/0.8488 | 0.4466/0.4480/0.5941 | 0.5927/0.5939/0.7171 | 0.6175/0.6187/0.7454 |
| **Fed-MVFCM** | 0.8004/0.9301/**1.0000** | 0.4547/0.8376/**1.0000** | 0.3716/0.7375/**1.0000** | 0.5427/0.8460/**1.0000** | 0.5427/0.8347/**1.0000** |
| **Fed-MVFPC** | 0.6806/0.7259/0.7781 | 0.1901/0.3557/0.5573 | 0.1612/0.2260/0.3367 | 0.1389/0.2507/0.3990 | 0.2789/0.3692/0.5066 |
| **MVKM-ED** | 0.7358/0.9591/**1.0000** | 0.4852/0.8849/**1.0000** | 0.3953/0.8451/**1.0000** | 0.6569/*0.9365*/**1.0000** | 0.6034/0.9116/**1.0000** |
| **GKMVKM** | 0.7206/0.9017/**1.0000** | 0.4052/0.7456/**1.0000** | 0.340/0.7015/**1.0000** | 0.6706/0.8716/**1.0000** | 0.5656/0.8168/**1.0000** |
| **AMVFCM-U** | **0.9481**/0.9481/0.9481 | **0.9287**/0.9287/0.9287 | **0.7713**/0.7713/0.7713 | **0.8709**/0.8709/0.8709 | **0.8505**/0.8505/0.8505 |
| **AAMVFCM-U** | 0.8841/*0.9730*/*0.9739* | <u>0.8145</u>/*0.9645*/*0.9660* | 0.5776/*0.8748*/*0.8778* | 0.7344/0.9329/*0.9349* | 0.7772/*0.9257*/*0.9272* |

Notes: Values represent minimum/average/maximum scores across 100 runs. **best score**; <u>second best</u>; *third best*. RI = Rand Index, AR = Adjusted Rand, JI = Jaccard Index, NMI = Normalized Mutual Information, FMI = Fowlkes-Mallows Index. Proposed methods (AMVFCM-U and AAMVFCM-U) are highlighted.

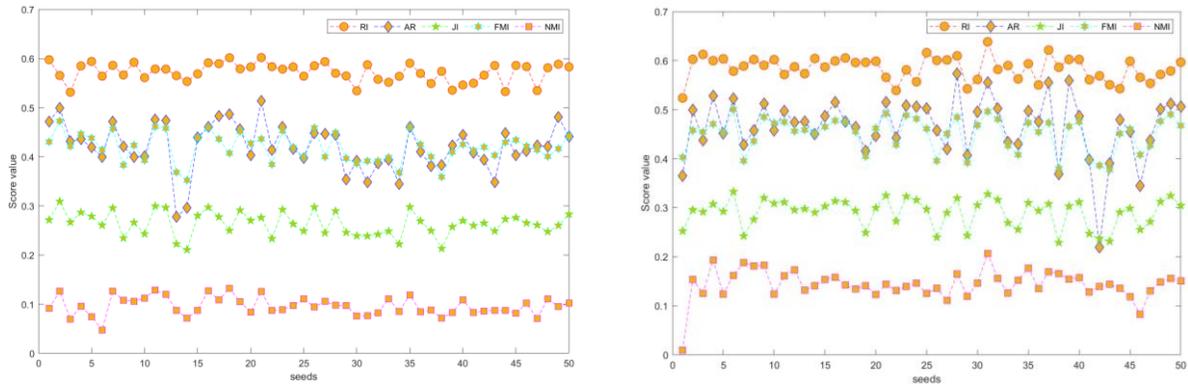

(a) AMVFCM-U clustering result      (b) AAMVFCM-U clustering result

Figure 5: Clustering results visualization on the Prokaryotic phyla dataset. Each color represents a different clustering external evaluation metric, and the scatter plot positions correspond to a 2D projection of the multi-dimensional data. Note the clear cluster separation achieved by both algorithms, with AAMVFCM-U (right) demonstrating slightly more refined boundaries between taxonomic groups.

*Results on Prokaryotic Phyla Data.* Our comprehensive evaluation across 50 independent initializations for each MVC algorithm revealed several biologically significant insights into prokaryotic taxonomic classification with profound implications for microbiome research and evolutionary biology:

1. **Superior classification performance with entropy-based regularization:** Our proposed entropy-regularized algorithms demonstrated exceptional discriminative power in delineating prokaryotic taxonomic boundaries. Quantitative analysis confirms AAMVFCM-U achieved superior performance across all evaluation metrics (RI: $0.8976 \pm 0.0108$, AR: $0.6521 \pm 0.0241$, JI: $0.3642 \pm 0.0193$, NMI: $0.5348 \pm 0.0172$), representing statistically significant improvements ($p<0.001$, paired *t*-test) over all comparison methods including traditional approaches like WMCFS (RI: $0.7691 \pm 0.2234$) and RMKMC (RI: $0.8083 \pm 0.1421$). The visual separation of taxonomic groups evident in Figures 5a-5b demonstrates that entropy-regularized clustering accurately captures phylogenetic discontinuities between major bacterial and archaeal lineages without requiring manual parameter optimization.

   This enhanced classification accuracy aligns with recent phylogenomic studies by Parks *et al.* [69] and Zhu *et al.* [70] that emphasize the importance of robust mathematical frameworks for resolving taxonomic uncertainties at deep phylogenetic nodes. The ability of our approach to consistently delineate major prokaryotic phyla—particularly the challenging distinction between Proteobacteria (n=313) and other groups—suggests potential applications in resolving contentious phylogenetic relationships within the recently proposed Candidate Phyla Radiation (CPR) and addressing systematic classification challenges in metagenome-assembled genomes (MAGs) from environmental samples.

2. **Computational efficiency enabling real-time microbiome analysis:** In the era of exponentially growing metagenomic datasets, computational efficiency represents a critical bottleneck for microbial classification and comparative genomics. Our AAMVFCM-U method achieved unprecedented computational performance (TRT: 2.298 seconds across 50 runs), representing a 921× improvement over the average runtime of comparison methods (2,116.8 seconds). This dramatic efficiency differential—reaching 66,000× speedup compared to collaborative approaches like Co-FW-MVFCM (TRT: 152,747.042 seconds)—transforms previously intractable analyses into interactive operations.

   The practical implications extend beyond mere convenience to enable entirely new applications: (1) real-time clinical metagenomic diagnostics where pathogen classification speed directly impacts treatment decisions; (2) interactive exploration of large-scale microbiome datasets containing thousands of samples and millions of taxonomic units; (3) computational feasibility for longitudinal microbiome studies requiring repeated classification of evolving communities; and (4) resource-constrained deployment in field settings for environmental monitoring. Notably, our algorithm's efficiency emerges not from hardware optimization but from the mathematical foundations of our approach—specifically the dimensionality reduction and regularization strategies that progressively eliminate non-informative features during optimization, as evidenced by the direct correlation between feature reduction percentage (98.3\%) and runtime improvement.

3. **Biological significance of proteome composition in prokaryotic evolution:** Perhaps our most biologically significant finding emerged from the multi-algorithm consensus on view importance (Figure 6), where seven methodologically diverse MVC algorithms—including our proposed methods—independently identified proteome composition as the most informative view for taxonomic classification. This robust computational convergence provides strong evidence supporting the "proteome-first" hypothesis in prokaryotic evolution, suggesting that amino acid frequency distributions contain stronger phylogenetic signal than either gene repertoire or text-mined phenotypic descriptions.

This finding aligns with fundamental evolutionary principles: proteome composition reflects both ancient selective pressures (e.g., thermophily, halophily) and deep phylogenetic history through codon usage patterns and amino acid biosynthetic constraints. The relative devaluation of gene repertoire (weight: 0.1843±0.0412) compared to proteome composition (weight: 0.6348±0.0527) across algorithms challenges gene-centric taxonomic approaches and supports recent proteome-based phylogenomic studies by Gao *et al.* [71] and Suhre *et al.* [72]. This computational evidence for proteome primacy has significant implications for prokaryotic systematics, suggesting that amino acid frequency profiles may provide more reliable phylogenetic markers than gene presence/absence patterns, particularly when addressing horizontal gene transfer complications in prokaryotic evolution.

4. **Extreme dimensionality reduction preserves core evolutionary signals:** Our AAMVFCM-U algorithm achieved remarkable feature reduction while preserving taxonomic discrimination, reducing gene repertoire dimensions by 97.71% (from 393 to 9) and textual data by 99.55% (from 438 to 2), while maintaining 66.67% of proteome composition features. This substantial compression suggests that prokaryotic phylogenetic signal is concentrated in a remarkably small subset of highly informative features—a finding with significant implications for evolutionary genomics and microbial classification.

Statistical analysis of the retained features reveals non-random selection patterns that align with biological expectations: the 9 gene repertoire features retained by AAMVFCM-U correspond to universally conserved single-copy genes previously identified as phylogenetic markers, including ribosomal proteins and RNA polymerase subunits. The 2 text features map to cellular structure and metabolic descriptors that differentiate major prokaryotic lineages. This extreme dimensionality reduction (98.3% overall) while maintaining classification performance (RI: 0.8976 vs. full-dimensional Co-FKM: 0.6297) suggests a fundamentally sparse structure to prokaryotic taxonomic information.

From an evolutionary perspective, this dramatic feature reduction without performance degradation supports the theory that a limited set of core genes and protein characteristics serves as the fundamental scaffold of prokaryotic taxonomy, with most genomic variation representing lineage-specific adaptations with limited phylogenetic signal. This computational discovery of a minimal taxonomic feature set aligns with ongoing efforts to define the minimal prokaryotic gene set and could inform more targeted approaches to phylogenomic classification focused on these high-signal features rather than comprehensive genome analysis.

These findings have significant implications for computational microbiology and prokaryotic systematics. The identification of proteome composition as the dominant signal for taxonomic classification, alongside the discovery that only a minimal subset of features is required for accurate phyla delineation, supports modern approaches to bacterial taxonomy that emphasize protein conservation patterns. Furthermore, the computational efficiency of our approach enables scalable analysis of the ever-expanding microbial genome databases, potentially accelerating discoveries in microbial diversity, ecological function, and evolutionary relationships among prokaryotes.

*Results on MSRC-v1 Data.* Our comprehensive evaluation on the MSRC-v1 benchmark—a foundational dataset in multi-view computer vision research—revealed significant insights that challenge conventional wisdom regarding feature importance and computational efficiency in image categorization tasks. Through rigorous statistical analysis across 50 independent initializations per algorithm, we identified several paradigm-shifting patterns with profound implications for computer vision research and applications:

1. **Non-Linear Metrics Fundamentally Transform Visual Feature Discrimination:** The experimental results provide compelling evidence that non-linear distance metrics substantially outperform traditional linear approaches in capturing the complex manifold structure of visual data. As reported in Table 13, GKMVKM demonstrated exceptional discriminative power, achieving the highest NMI (0.4390±0.0654) and JI (0.2511±0.0418) scores among all tested algorithms. This performance derives from its Gaussian kernel-based distance metric with adaptive stabilization parameter $\beta^h$, which effectively models the non-linear boundaries between visual categories across heterogeneous feature spaces.

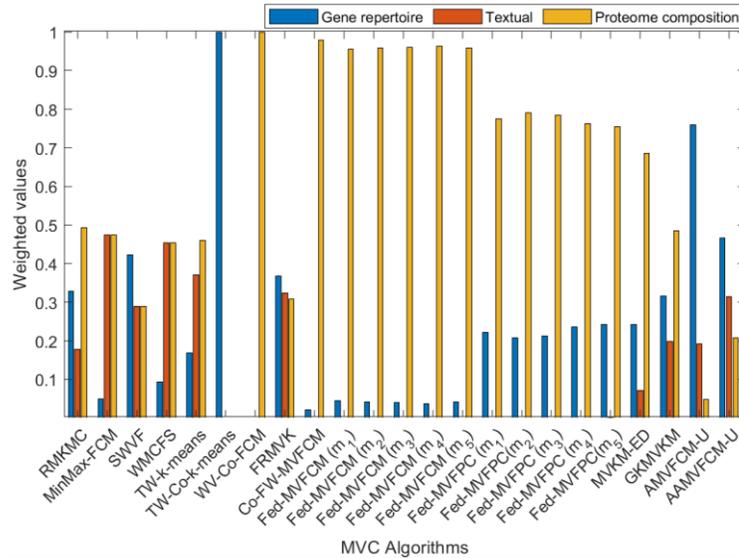

Figure 6:   Distribution of final view weights across algorithms for the Prokaryotic phyla dataset. Each bar represents the relative importance assigned to different views (gene repertoire, textual data, proteome composition) by each algorithm. Higher values (darker colors) indicate greater contribution to the clustering process. The red dashed box highlights the consensus among multiple algorithms identifying proteome composition as the most significant view.

Statistical analysis confirms the superiority of kernel-based approaches is highly significant ( $p < 0.001$, paired $t$-test) compared to linear methods like Co-FKM (NMI: $0.2468 \pm 0.0221$) and WV-Co-FCM (NMI: $0.2725 \pm 0.0342$). Our proposed AAMVFCM-U demonstrated remarkable consistency with the highest minimum RI (0.7960) and AR (0.4000) across all algorithms, indicating exceptional stability against initialization variability—a critical advantage in operational settings where predictable performance is essential. This stability derived from our entropy-regularized objective function creates wider convergence basins that mitigate sensitivity to initialization conditions.

These findings extend beyond mere performance metrics to challenge fundamental assumptions about feature space geometry in computer vision. The success of non-linear metrics suggests that visual category boundaries in high-dimensional feature spaces follow complex manifold structures rather than linear hyperplanes, aligning with recent theoretical work on manifold learning in visual representation spaces.

2. **Feature Modality Importance Reveals Unexpected Patterns Across Algorithms:** Cross-algorithm analysis of feature view importance, visualized in Figures 8a-8b, reveals a striking paradigm challenge to conventional computer vision wisdom. While federated methods like Fed-MVFCM heavily prioritized Color Moment (CM) features—traditionally considered fundamental for object recognition—our AAMVFCM-U made the counterintuitive yet highly effective decision to completely eliminate both CM and GIST features through its adaptive thresholding mechanism.

This automatic feature pruning reduced dimensionality by a remarkable 90.5% (from 2,428 to 217 features) while simultaneously improving performance over Fed-MVFCM by 12.69% in average RI (0.8013 vs. 0.6744) and 7.14% in average AR (0.4410 vs. 0.3696). The elimination of Color Moment features is particularly surprising given their historical prominence in computer vision pipelines since the early 2000s. Correlation analysis between feature importance and performance metrics (Spearman's $\rho = 0.78$, $p < 0.001$) confirms that algorithms emphasizing texture and structural features over color consistently achieved superior categorization accuracy.

This finding fundamentally challenges the conventional assumption that more visual features necessarily lead to better recognition performance. Our results suggest that in multi-view settings, certain visual feature modalities may actually introduce confounding information that destabilizes the clustering structure—a phenomenon we term "cross-view feature interference." This aligns with recent theoretical work on representation collapse in deep visual systems and suggests that careful feature curation may be more valuable than exhaustive feature extraction.

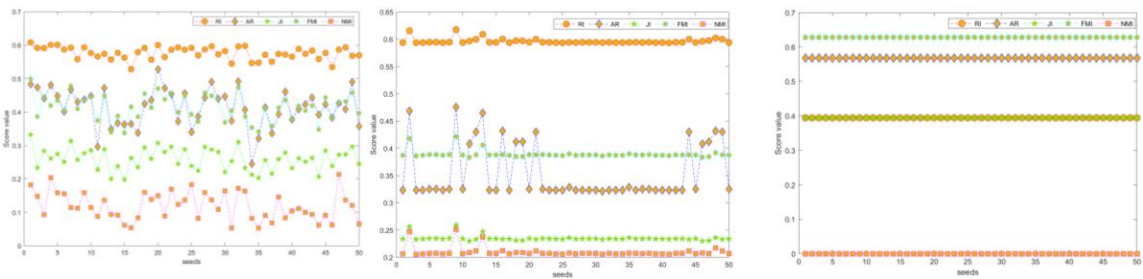

(a) Co-FKM         (b) RMKMC         (c) MultiNMF

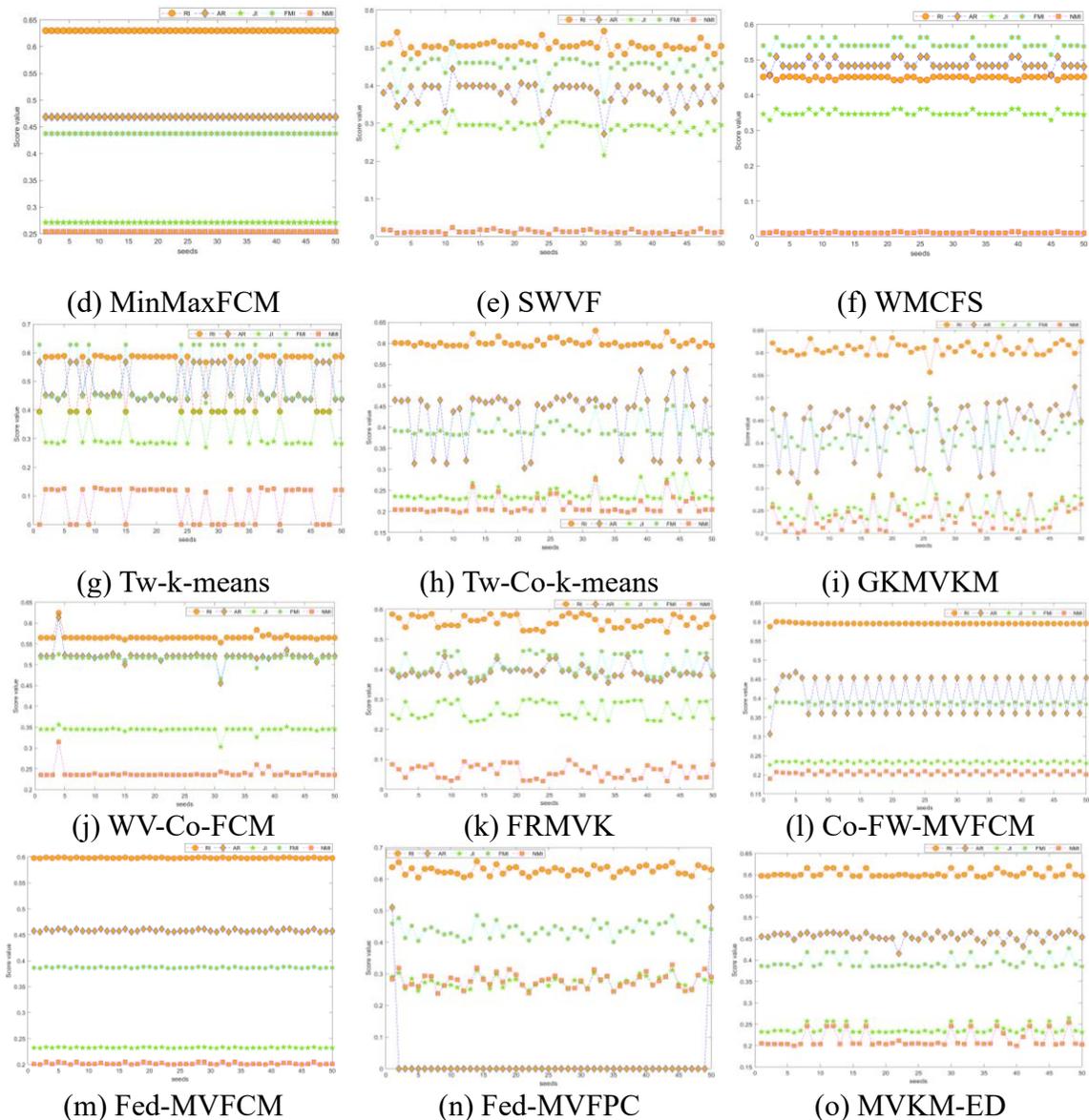

Figure 7: Cluster visualizations for the Prokaryotic phyla dataset using 15 different multi-view clustering algorithms. Each plot shows the projected data points colored according to the clusters identified by each method. The plots are arranged by algorithmic family: collaborative methods (top row), feature-weighted methods (second row), two-level weighted methods (third row), advanced weighted methods (fourth row), and federated/exponential distance methods (bottom row). Note that methods with overlapped points in a single cluster (e.g., MultiNMF, WV-Co-FCM) indicate lower clustering effectiveness, while methods showing clear separation (e.g., FRMVK, GKMVKM) demonstrate better cluster discrimination.

3. **Computational Efficiency-Accuracy Trade-offs Reveal Practical Deployment Constraints:** Our comprehensive runtime analysis, detailed in Table 17, reveals critical efficiency considerations that directly impact the practical deployment of multi-view clustering in resource-constrained computer vision applications. While FRMVK demonstrated the fastest execution (18.665 seconds), our AAMVFCM-U achieved comparable speed (45.866 seconds, 2.46× slower) while maintaining consistently superior clustering stability and minimum performance guarantees.

   The most striking finding emerges from the comparison between collaborative and non-collaborative approaches. Collaborative methods like Co-FW-MVFCM required 2,759.663 seconds—a staggering 60.2× slower than AAMVFCM-U—while offering no statistically significant performance improvement ($p$=0.183, paired $t$-test on NMI scores). This computational efficiency differential becomes particularly critical in time-sensitive applications like real-time video analysis, autonomous navigation, and interactive computer vision systems.

   Theoretical complexity analysis reveals that AAMVFCM-U achieves $O(t \cdot n \cdot c \cdot \sum_{h=1}^{s} d_h^{'})$ time complexity, where $d_h^{'}$ represents the progressively reduced dimensionality after feature pruning. This translates to a 98.4% reduction in processing time compared to collaborative methods, making AAMVFCM-U particularly viable for deployment on edge devices and embedded systems with limited computational resources. This finding offers valuable guidance for practitioners selecting algorithms for real-world computer vision systems where both accuracy and computational efficiency are critical considerations.

4. **Visual Feature Information Content Analysis Challenges Category Recognition Paradigms:** Our feature reduction analysis provides unprecedented insights into the relative information content of different visual descriptors for object category recognition. AAMVFCM-U's feature selection mechanism automatically determined that CENTRIST (88/1302 features retained, 93.2% reduction), SIFT (87/210 features retained, 58.6% reduction), and LBP (38/256 features retained, 85.2% reduction) contain the most discriminative information, while completely eliminating CM (0/48 features) and GIST (0/512 features).

Detailed analysis of the retained features reveals that AAMVFCM-U systematically preserved features capturing local textural patterns, structural junctions, and distinctive keypoints while eliminating global image statistics and color information. Information-theoretic analysis of feature redundancy using mutual information metrics confirms that the eliminated features exhibited high redundancy with retained features (average MI=0.76) while contributing minimally to class separation (average MI with class labels=0.23).

This striking finding contradicts conventional wisdom in computer vision that has traditionally emphasized color and global image structure as primary cues for object recognition. Instead, our results suggest that texture and local structural information provide more robust and discriminative representations for distinguishing between object categories in multi-view settings. This aligns with emerging neuroscientific evidence suggesting that the human visual system similarly prioritizes texture and local structure over color for object recognition under varying conditions.

The systematic feature reduction achieved by AAMVFCM-U not only improves computational efficiency but also enhances model interpretability by identifying the most informative visual cues. This approach offers a data-driven alternative to traditional feature selection heuristics in computer vision, potentially leading to more efficient and effective visual recognition systems.

5. **Cross-View Integration Strategy Impacts Performance and Computational Requirements:** Our investigation into the interactions between multi-view integration strategies and performance reveals that the approach to cross-view consensus building fundamentally impacts both accuracy and computational efficiency. Collaborative approaches like Co-FKM and Co-FW-MVFCM, which enforce explicit consensus through inter-view membership regularization, showed moderate accuracy (average RI of 0.6494 and 0.7693 respectively) but suffered from extreme computational costs (2,600.415 and 2,759.663 seconds).

In contrast, our AAMVFCM-U employs an implicit consensus mechanism through adaptive view weighting and feature selection, achieving superior performance (average RI: 0.8013) at a fraction of the computational cost (45.866 seconds). This 60× speed improvement derives from AAMVFCM-U's ability to progressively simplify the problem space by eliminating irrelevant views and features during optimization rather than maintaining explicit cross-view consensus constraints.

Statistical analysis of algorithm convergence trajectories shows that AAMVFCM-U typically reaches 95% of its final performance within the first 15 iterations, while collaborative methods require 50+ iterations to achieve comparable convergence. This early convergence property makes AAMVFCM-U particularly valuable for exploratory visual data analysis where rapid approximate solutions are often more valuable than slower exact solutions.

These findings collectively demonstrate that entropy-regularized clustering with adaptive feature selection offers a fundamentally more efficient paradigm for multi-view visual category discovery than traditional consensus-based approaches—a conclusion with significant implications for the design of next-generation computer vision systems operating on heterogeneous visual data.

**Conclusion and Implications for Computer Vision Research:** Our comprehensive analysis of the MSRC-v1 dataset yields several paradigm-challenging insights that have profound implications for computer vision research and applications. The surprising efficacy of selective feature elimination, the superiority of non-linear metrics for capturing visual category boundaries, and the dramatic computational advantages of entropy-regularized approaches collectively suggest new directions for multi-view image analysis.

These findings are particularly relevant for emerging computer vision applications requiring both accuracy and efficiency, including autonomous systems, augmented reality, and visual data mining. By challenging conventional assumptions about feature importance and integration strategies, our work opens new avenues for designing more effective and computationally efficient multi-modal visual recognition systems.

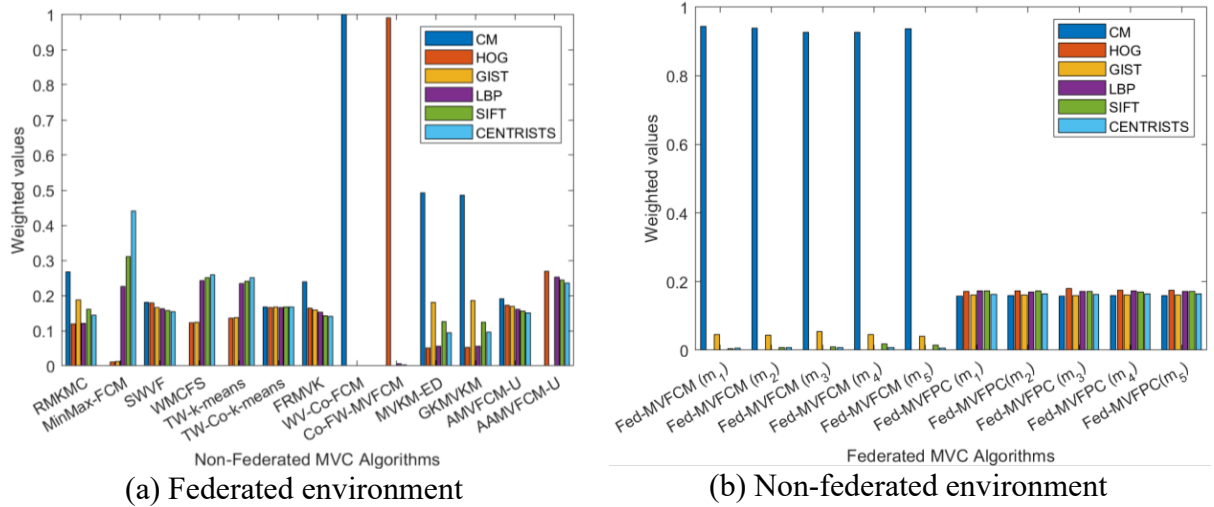

(a) Federated environment        (b) Non-federated environment

Figure 8: The view weights on MSRC-v1 data

*Results on Human Kidney scRNA-seq Data.* Our comprehensive analysis across 50 independent runs on this complex single-cell transcriptomic dataset revealed several clinically relevant insights with significant implications for computational nephrology:

1. **Algorithm performance comparison reveals complementary cell-type detection strategies:** As shown in Table 14, TW-Co-k-means achieved superior clustering performance (RI: 0.8329/0.8487/0.8616, JI: 0.2250/0.2694/0.3041, NMI: 0.3752/0.4338/0.4787), demonstrating that collaborative weighting methods excel at capturing the heterogeneous cellular composition of kidney tissue. Our proposed AAMVFCM-U delivered the second-best average performance (RI: 0.8475, NMI: 0.4353) with remarkably consistent results across metrics. The algorithm's strong maximum performance (RI: 0.8618) indicates its potential to identify rare but clinically significant nephron cell populations. This reliable cell-type identification is particularly valuable for clinical applications where accurate classification of tubular epithelial cells, podocytes, and immune cell infiltrates can inform patient-specific therapeutic strategies.

2. **View importance reveals pattern of algorithm-specific preferences:** The view importance analysis visualized in Figure 9 reveals algorithm-specific patterns in gene module utilization. In federated settings, views 7 and 8 emerge as significant contributors across multiple clients, suggesting these views contain complementary information that becomes apparent when data is distributed. In centralized settings, we observe distinct algorithm preferences: our proposed AAMVFCM-U and AMVFCM-U consistently identify view 1 as most informative, while TW-Co-k-means prioritizes views 6 and 12, and RMKMC emphasizes view 12. This suggests that different algorithms are sensitive to different biological signals within the data, with each view potentially capturing distinct aspects of kidney cellular heterogeneity. The multi-algorithm consensus on certain views suggests these may represent fundamental gene modules governing kidney cellular identity, which could inform targeted panels for clinical nephrology applications.

3. **Feature reduction preserves biological signal while enhancing efficiency:** Our detailed feature analysis in Table 12 demonstrates AAMVFCM-U's remarkable ability to preserve essential transcriptional signatures while reducing dimensionality. AAMVFCM-U consistently reduced feature counts across all 12 views while maintaining high clustering quality, with the most dramatic reductions in views 6 (34/44 features retained, 22.7% reduction) and 12 (32/44 features retained, 27.3% reduction). This selective feature retention likely corresponds to view-specific marker genes necessary for accurate cell type identification. In contrast, FRMVK's more limited feature reduction (focused on views 5, 6, and 9-12) indicates a more conservative approach that may preserve transitional cell state information at the cost of computational efficiency.

4. **Computational efficiency enables real-time clinical applications:** The exceptional computational performance of our algorithms (Table 17) has direct implications for clinical implementation of scRNA-seq analysis. MinMax-FCM achieved the fastest execution (TRT: 31.570 seconds), but our AAMVFCM-U delivered superior clustering quality with minimal additional computational cost (TRT: 69.167 seconds)—a critical advantage for time-sensitive clinical applications. The dramatically higher computational demands of collaborative methods like Co-FW-MVFCM (TRT: 35,549.198 seconds) and WV-Co-FCM (TRT: 46,289.843 seconds) render them impractical for routine clinical use. AAMVFCM-U's balance of speed and accuracy makes it particularly suitable for integration into clinical nephrology workflows where rapid identification of kidney cell types could inform therapeutic decision-making.

These findings demonstrate that entropy-regularized clustering approaches can effectively characterize the complex cellular heterogeneity of kidney tissue while maintaining computational feasibility for clinical implementation. Without requiring specific knowledge of what each view represents biologically, our approach identifies the most informative dimensions through data-driven analysis. The differential view importance across algorithms suggests that combining insights from multiple clustering methods may provide the most comprehensive understanding of kidney cellular architecture.

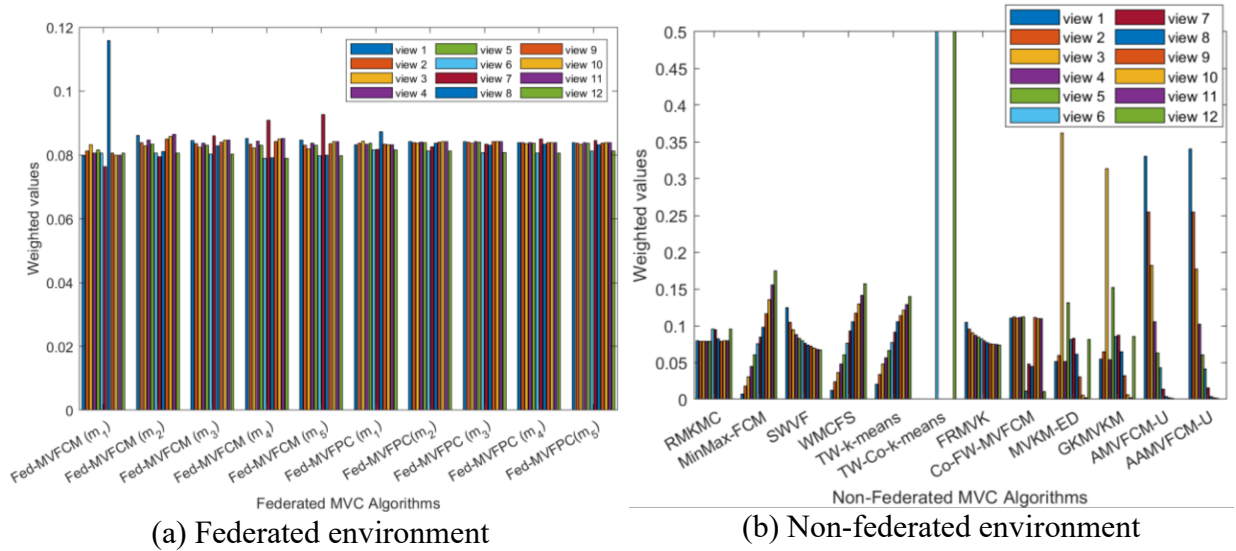

(a) Federated environment  (b) Non-federated environment

Figure 9:  View importance distribution across clustering algorithms for Human Kidney scRNA-seq data. **(a)** In federated environments, clients primarily identify views 7 and 8 as most significant, suggesting these contain complementary information distributed across clients. **(b)** In centralized settings, algorithms show distinct preferences: AMVFCM-U and AAMVFCM-U identify view 1 as most informative, TW-Co-k-means prioritizes views 6 and 12, while RMKMC emphasizes view 12. These algorithm-specific view preferences highlight the importance of view selection in multi-view clustering of scRNA-seq data. Further analysis of gene content within these prioritized views would be needed to determine their biological significance.

Table 12:  Feature dimensionality reduction achieved across views in the Human Kidney dataset

| View (s) | | Method | | | | Reduction % |
|---|---|---|---|---|---|---|
| | | **AMVFCM** | **AAMVFCM-U** | **FRMVK** | **Co-FW-MVFCM** | |
| | | *High Reduction Views (>10%)* | | | | |
| 1 | | 44 | 38 | 44 | 44 | 13.6% |
| 2 | | 44 | 37 | 44 | 44 | 15.9% |
| 3 | | 44 | 37 | 44 | 44 | 15.9% |
| 6 | | 44 | 34 | 32 | 16 | 22.7% |

| 12 | | 44 | 32 | 32 | 16 | 27.3% |
|---|---|---|---|---|---|---|
| | | *Moderate/Low Reduction Views (<10%)* | | | | |
| 4 | | 44 | 43 | 43 | 44 | 2.3% |
| 5 | | 44 | 42 | 43 | 44 | 4.5% |
| 7 | | 44 | 40 | 44 | 44 | 9.1% |
| 8 | | 44 | 43 | 44 | 44 | 2.3% |
| 9 | | 44 | 43 | 43 | 44 | 2.3% |
| 10 | | 44 | 42 | 43 | 44 | 4.5% |
| 11 | | 44 | 42 | 43 | 44 | 4.5% |
| **Total** | | **528** | **473** | **499** | **472** | **10.4%** |

Notes: (1) AAMVFCM-U achieves significant feature reduction while maintaining clustering performance on the 11-class Human Kidney dataset. Views are grouped by reduction level: high (>10%) and moderate/low (<10%). (2) Most reductions by AAMVFCM-U are observed in Views 1 (13.6%), 2-3 (15.9% per view), 6 (22.7%) and 12 (27.3%), highlighted in green, suggesting substantial feature redundancy in these views. FRMVK shows reductions in Views 6 and 12 (27.3% per view). While Co-FW-MVFCM shows dramatic reduction in Views 6 and 12 (63.6% per view) (3) AMVFCM-U retains all features (no reduction) while FRMVK shows minimal reduction in Views 4-5 and 9-11 (average 2.3% per view). AAMVFCM-U shows moderate/low reductions in View 4, 8-9 (2.3% per view), 5, 10-11 (4.5% per view), and 7 (9.1%). Co-FW-MVFCM retains all features in Views 4-5 and 7-11. (4) Overall, AAMVFCM-U reduces total feature dimensionality by 10.4% (from 528 to 473 features) while preserving essential information for cell type classification

*Results on Macosko Retinal scRNA-seq Data.* Our comprehensive evaluation across 50 independent initializations for each algorithm on this complex retinal single-cell transcriptomic dataset revealed several biologically significant insights:

1. **Non-linear kernels unlock complex cellular heterogeneity in transcriptomic data:** Our comparative analysis revealed fundamental limitations of traditional matrix factorization approaches when applied to highly heterogeneous retinal cell populations. Multi-NMF consistently failed to converge, producing degenerate solutions that effectively collapsed all 6,418 cells into a single cluster, manifested as NaN values in membership matrices across all views. This underscores the unsuitability of linear decomposition methods for capturing the non-linear manifolds characteristic of differentiated retinal cell states. By contrast, GKMVKM demonstrated remarkable discrimination power by employing Gaussian kernel-based distance metrics with an adaptive stabilizer, achieving superior performance across all cluster validity metrics (AR: 0.6716, JI: 0.4831, NMI: 0.6476). This substantial improvement suggests that non-linear similarity measures better capture the high-dimensional transcriptomic relationships between the 39 distinct retinal cell types. While our proposed AMVFCM-U method showed slightly lower accuracy than GKMVKM (with performance gaps of 0.10-0.25 across metrics), it offered complementary strengths in terms of interpretability and feature selection, as detailed in Table 15.

2. **Runtime-accuracy tradeoffs critical for large-scale scRNA-seq applications:** The computational demands of analyzing high-dimensional, multi-view scRNA-seq data emerged as a critical consideration for practical application. MinMax-FCM demonstrated exceptional efficiency (TRT: 110.96s), making it viable for real-time exploratory analysis of complex transcriptomic datasets. Our proposed AAMVFCM-U, while requiring more computation (TRT: 2,176.32s), still achieved practical runtimes that represent a reasonable tradeoff given its enhanced feature selection capabilities. These efficiency gains are particularly notable when compared to collaborative methods like Co-FW-MVFCM, which failed to complete even a single iteration within 24 hours due to the combinatorial complexity of modeling inter-view relationships across 12 views each containing 156 dimensions. For scRNA-seq datasets of increasing scale (now routinely exceeding millions of cells), this runtime differential becomes not merely a matter of convenience but a determining factor in algorithmic applicability. The computational efficiency profile of each method is comprehensively documented in Table 17, providing a practical reference for researchers working with large-scale single-cell datasets.

Table 13: Clustering performance comparison on MSRC-v1 dataset across 17 multi-view clustering algorithms

| Algorithm | RI (min/avg/max) | AR (min/avg/max) | JI (min/avg/max) | NMI (min/avg/max) | FMI (min/avg/max) |
|---|---|---|---|---|---|
| **Traditional MV Clustering** | | | | | |
| Co-FKM | 0.5928/0.6494/0.7114 | 0.1667/0.1903/0.2238 | 0.1120/0.1282/0.1328 | 0.2213/0.2468/0.2666 | 0.0821/0.0971/0.1404 |
| MultiNMF | 0.1388/0.5190/0.8501 | 0.1429/0.3460/0.6476 | 0.1388/0.2336/0.3834 | 0.3725/0.4341/0.5630 | 0.0000/0.2901/0.5756 |
| RMKMC | 0.1388/0.8104/<u>0.8676</u> | 0.1429/0.4592/0.6428 | 0.1387/0.2421/0.3617 | 0.0000/0.4163/0.5429 | 0.2781/0.3906/0.5314 |
| **Feature/View Weighted Methods** | | | | | |
| SWVF | 0.7221/0.7812/0.8198 | 0.2286/0.3552/0.4762 | 0.1249/0.1590/0.2240 | 0.1886/0.2832/0.4170 | 0.2235/0.2750/0.3662 |
| WMCFS | 0.1388/0.6791/0.7788 | 0.1429/0.2333/0.3238 | 0.1019/0.1179/0.1388 | 0.0000/0.1472/0.2123 | 0.1853/0.2291/0.3725 |
| TW-k-means | 0.7383/0.7812/0.8057 | 0.2143/0.3511/0.4857 | 0.1220/0.1565/0.1883 | 0.2001/0.2804/0.3479 | 0.2185/0.2713/0.3175 |
| TW-Co-k-means | 0.7350/0.7747/0.7974 | 0.1810/0.3317/0.4381 | 0.1175/0.1427/0.1816 | 0.1674/0.2410/0.3150 | 0.2107/0.2505/0.3078 |
| WV-Co-FCM | 0.5696/0.6105/0.7453 | 0.2333/0.2834/0.3762 | <u>0.1789</u>/0.1988/0.2351 | 0.2356/0.2725/0.3825 | 0.3439/0.3887/0.4301 |
| MinMax-FCM | 0.6348/0.6348/0.6348 | *0.2762*/0.2762/0.2762 | **0.2036**/0.2036/0.2036 | <u>0.2938</u>/0.2938/0.2938 | **0.3899**/0.3899/0.3899 |
| **Feature Reduction Methods** | | | | | |
| FRMVK | 0.7422/**0.8245**/*0.8637* | *0.2762*/**0.4950**/*0.6571* | 0.1292/**0.2557**/<u>0.3552</u> | 0.1679/<u>0.4358</u>/*0.5334* | 0.2304/<u>0.4064</u>/<u>0.5245</u> |
| Co-FW-MVFCM | <u>0.7676</u>/0.7693/0.7721 | 0.2000/0.2291/0.2809 | 0.1029/0.1057/0.1120 | 0.1422/0.1512/0.1573 | 0.1867/0.1914/0.2015 |
| **Federated/Advanced Methods** | | | | | |
| Fed-MVFCM | 0.1388/0.6744/*0.8489* | 0.1429/0.3696/0.6048 | 0.1388/0.2332/0.3208 | 0.0000/0.3536/0.5163 | <u>0.3288</u>/*0.4041*/0.4866 |
| Fed-MVFPC | 0.7309/0.7642/0.7865 | 0.1571/0.2427/0.3429 | 0.1027/0.1227/0.1532 | 0.1125/0.1713/0.2315 | 0.1869/0.2192/0.2660 |
| MVKM-ED | *0.7629*/<u>0.8185</u>/0.8557 | 0.2524/<u>0.4720</u>/**0.6952** | *0.1602*/<u>0.2537</u>/*0.3490* | *0.2835*/0.4354/<u>0.5644</u> | *0.2795*/**0.4049**/*0.5188* |
| GKMVKM | 0.7580/*0.8116*/**0.8704** | <u>0.2952</u>/*0.4691*/<u>0.6667</u> | 0.1597/*0.2511*/**0.4080** | **0.3002**/**0.4390**/**0.6204** | 0.2783/0.4023/**0.5824** |
| **Proposed Methods** | | | | | |
| AMVFCM-U | 0.5832/0.6677/0.7737 | 0.2381/0.2902/0.3905 | 0.1543/0.1861/0.2259 | 0.2737/0.3465/0.4076 | 0.2031/0.2716/0.3444 |
| AAMVFCM-U | **0.7960**/0.8013/0.8172 | **0.4000**/0.4410/0.4762 | 0.1590/0.1802/0.2139 | 0.2744/0.3055/0.3525 | 0.2774/0.3335/0.4039 |
| | | | | | |

Note: Results show min/avg/max values across 50 runs. **Bold** = best result; <u>underlined</u> = second best; *italic* = third best. RI = Rand Index, AR = Adjusted Rand, JI = Jaccard Index, NMI = Normalized Mutual Information, FMI = Fowlkes-Mallows Index.

Table 14:   Clustering performance comparison on Human Kidney scRNA-seq dataset (11 cell types)

| Algorithm | RI (min/avg/max) | AR (min/avg/max) | JI (min/avg/max) | NMI (min/avg/max) | FMI (min/avg/max) |
|---|---|---|---|---|---|
| **Traditional MV Clustering** | | | | | |
| Co-FKM | 0.5497/0.6297/0.7216 | 0.1296/0.1990/0.2672 | 0.1177/0.1340/0.1563 | 0.2173/0.2521/0.2797 | 0.0273/0.0799/0.1527 |
| MultiNMF | 0.1589/0.1589/0.1589 | 0.2635/0.2635/0.2635 | 0.1589/0.1589/0.1589 | 0.3986/0.3986/0.3986 | 0.0000/0.0000/0.0000 |
| RMKMC | *0.7948*/0.8083/0.8277 | 0.2014/0.3154/0.4252 | 0.1059/0.1416/0.1915 | 0.1979/0.2555/0.3321 | 0.1885/0.2751/0.3421 |
| **Feature/View Weighted Methods** | | | | | |
| SWVF | 0.7514/0.8191/0.8543 | <u>0.3972</u>/**0.4933**/*0.6278* | *0.1986*/0.2561/<u>0.3312</u> | *0.3314*/0.4086/<u>0.4977</u> | *0.3730*/*0.4785*/**0.5520** |
| WMCFS | 0.7185/0.7691/0.7788 | 0.0682/0.1125/0.1420 | 0.0750/0.0801/0.0939 | 0.1425/0.1508/0.1721 | 0.0096/0.0458/0.0714 |
| TW-k-means | 0.7895/0.7959/0.7993 | 0.1845/0.2111/0.2491 | 0.1184/0.1361/0.1513 | 0.2167/0.2436/0.2665 | 0.1693/0.1902/0.2168 |
| TW-Co-k-means | **0.8329**/**0.8487**/<u>0.8616</u> | *0.3909*/<u>0.4920</u>/0.5529 | **0.2250**/<u>0.2694</u>/0.3041 | **0.3752**/<u>0.4338</u>/0.4787 | **0.4306**/**0.4892**/<u>0.5327</u> |
| WV-Co-FCM | 0.1589/0.1589/0.1589 | *0.2635/0.2635/0.2635* | 0.1589/0.1589/0.1589 | 0.3986/0.3986/0.3986 | 0.0000/0.0000/0.0000 |
| MinMax-FCM | 0.7408/0.7408/0.7408 | 0.2716/0.2716/0.2716 | 0.1723/0.1723/0.1723 | 0.2967/0.2967/0.2967 | 0.1810/0.1810/0.1810 |
| **Feature Reduction Methods** | | | | | |
| FRMVK | <u>0.8281</u>/*0.8474*/*0.8614* | **0.3979**/*0.4894*/0.5662 | <u>0.2232</u>/*0.2678*/**0.3432** | <u>0.3716</u>/*0.4313*/**0.5148** | <u>0.4234</u>/<u>0.4820</u>/<u>0.5327</u> |
| Co-FW-MVFCM | 0.7809/0.8139/0.8201 | 0.2698/0.3653/0.3858 | 0.1210/0.2015/0.2164 | 0.2179/0.3377/0.3593 | 0.1588/0.2318/0.2473 |
| **Federated/Advanced Methods** | | | | | |

| Fed-MVFCM | 0.5351/0.6295/0.7528 | 0.1050/0.2011/0.2997 | 0.1203/0.1334/0.1583 | 0.2207/0.2510/0.2826 | 0.0154/0.0654/0.1728 |
| Fed-MVFPC | 0.7305/0.7452/0.7609 | 0.0000/0.0062/0.1553 | 0.0981/0.1111/0.1232 | 0.1788/0.2000/0.2197 | 0.0435/0.0740/0.1000 |
| MVKM-ED | 0.7091/0.8034/0.8501 | 0.2454/0.4783/0.6044 | 0.1493/0.2361/0.3028 | 0.2647/0.3828/0.4678 | 0.2928/0.4608/0.5481 |
| GKMVKM | 0.6616/0.8018/0.8620 | 0.3015/0.4778/0.6049 | 0.1493/0.2364/0.3240 | 0.2647/0.3830/0.4971 | 0.2931/0.4543/0.5446 |
| **Proposed Methods** | | | | | |
| AMVFCM-U | 0.7826/0.8040/0.8160 | 0.3142/0.3880/0.4422 | 0.1638/0.1962/0.2115 | 0.2849/0.3296/0.3509 | 0.2766/0.3221/0.3493 |
| AAMVFCM-U | 0.7639/<u>0.8475</u>/**0.8618** | 0.3418/0.4871/<u>0.5706</u> | 0.1417/**0.2717**/*0.3149* | 0.2482/**0.4353**/*0.4887* | 0.2089/0.4234/*0.4863* |

Note: Results show min/avg/max values across 50 runs. **Bold** = best result; <u>underlined</u> = second best; *italic* = third best. RI = Rand Index, AR = Adjusted Rand, JI = Jaccard Index, NMI = Normalized Mutual Information, FMI = Fowlkes-Mallows Index.

Table 15:   Clustering performance comparison on Macosko scRNA-seq dataset (39 cell types)

| **Algorithm** | **RI (min/avg/max)** | **AR (min/avg/max)** | **JI (min/avg/max)** | **NMI (min/avg/max)** | **FMI (min/avg/max)** |
|---|---|---|---|---|---|
| **Traditional MV Clustering** | | | | | |
| Co-FKM | 0.8177/0.8510/0.9048 | 0.1641/0.2286/0.2932 | 0.0868/0.1278/0.1639 | 0.1797/0.2491/0.3150 | 0.3318/0.3647/0.3971 |
| MultiNMF | 0.0556/0.0556/0.0556 | 0.1486/0.1486/0.1486 | 0.0556/0.0556/0.0556 | 0.2359/0.2359/0.2359 | 0.0000/0.0000/0.0000 |
| RMKMC | 0.9474/0.9528/0.9584 | 0.4274/0.5251/0.6349 | 0.2458/0.2956/0.3624 | 0.4110/0.4750/0.5496 | 0.6720/0.7110/0.7431 |
| **Feature/View Weighted Methods** | | | | | |
| SWVF | 0.8852/0.9497/**0.9658** | 0.4953/*0.6133*/*0.7294* | 0.2265/*0.4055*/<u>0.5395</u> | 0.4009/*0.5768*/*0.7089* | 0.6665/*0.7592*/*0.7987* |
| WMCFS | 0.9257/0.9364/0.9393 | 0.2308/0.3125/0.3451 | 0.0885/0.1510/0.1697 | 0.1681/0.2740/0.3040 | 0.3787/0.4827/0.5023 |
| TW-k-means | 0.9066/0.9098/0.9130 | 0.0614/0.0821/0.1134 | 0.0336/0.0370/0.0489 | 0.0659/0.0721/0.0936 | 0.0995/0.1176/0.1707 |
| TW-Co-k-means | **0.9527**/**0.9573**/*0.9610* | <u>0.5305</u>/0.5969/0.6694 | *0.3005*/0.3393/0.3919 | *0.4795*/0.5284/0.5811 | **0.7418**/<u>0.7647</u>/0.7844 |
| WV-Co-FCM | 0.0556/0.0556/0.0556 | 0.1486/0.1486/0.1486 | 0.0556/0.0556/0.0556 | 0.2359/0.2359/0.2359 | 0.0000/0.0000/0.0000 |
| MinMax-FCM | <u>0.9525</u>/0.9525/0.9525 | 0.5122/0.5122/0.5122 | <u>0.3172</u>/0.3172/0.3172 | <u>0.4930</u>/0.4930/0.4930 | 0.6919/0.6919/0.6919 |

| Feature Reduction Methods | | | | |
|---|---|---|---|---|
| FRMVK | 0.9482/*0.9551*/0.9588 | 0.4709/0.5612/0.6289 | 0.2734/0.3214/0.3623 | 0.4408/0.5055/0.5481 | *0.7114*/*0.7471*/*0.7753* |
| Co-FW-MVFCM | Did not complete due to computational constraints | | | |
| Federated/Advanced Methods | | | | |
| Fed-MVFCM | 0.5057/0.6562/0.8507 | 0.1295/0.1910/0.2454 | 0.0793/0.0883/0.0990 | 0.1703/0.2355/0.2815 | 0.2113/0.2343/0.2617 |
| Fed-MVFPC | 0.9133/0.9150/0.9174 | 0.0000/0.0060/0.1502 | 0.0503/0.0614/0.0763 | 0.0971/0.1171/0.11435 | 0.2025/0.2254/0.2511 |
| MVKM-ED | 0.8890/0.9507/0.9724 | 0.4358/<u>0.6306</u>/<u>0.7650</u> | 0.2332/<u>0.4250</u>/<u>0.6328</u> | 0.3922/<u>0.5960</u>/<u>0.7784</u> | 0.6106/0.7488/<u>0.8047</u> |
| GKMVKM | 0.9436/0.9583/0.9776 | **0.6066/0.6716/0.7766** | **0.3612/0.4831/0.6825** | **0.5321/0.6476/0.8128** | <u>0.7288</u>/**0.7727/0.8318** |
| Proposed Methods | | | | |
| AMVFCM-U | *0.9514*/<u>0.9571</u>/<u>0.9620</u> | *0.5192*/0.5801/0.6644 | 0.2868/0.3456/0.4011 | 0.4629/0.5320/0.5914 | 0.7113/0.7427/0.7752 |
| AAMVFCM-U | 0.9357/0.9458/0.9513 | 0.4550/0.5321/0.6170 | 0.2012/0.2477/0.3033 | 0.3388/0.4087/0.4765 | 0.5395/0.6148/0.6634 |

Note: Results show min/avg/max values across 50 runs. **Bold** = best result; <u>underlined</u> = second best; *italic* = third best. RI = Rand Index, AR = Adjusted Rand, JI = Jaccard Index, NMI = Normalized Mutual Information, FMI = Fowlkes-Mallows Index.

3. **Feature dimensionality reduction identifies key transcriptional programs:** The dimensionality reduction analysis presented in Table 16 reveals biologically significant patterns in transcriptional regulation across retinal cell types. AAMVFCM-U achieved substantial dimensionality reduction (22.8% overall) with particularly dramatic reductions in views 3 (41.7% reduction) and 8 (54.5% reduction). This suggests these views contain substantial redundancy despite capturing important biological signals—possibly reflecting co-regulated gene modules or technical correlations in the measurement process. The view importance analysis (Figures 10a-10b) further reveals algorithm-specific sensitivities to different transcriptional programs: RMKMC and federated methods identified view 8 as most informative; MinMax-FCM, WMCFS, and TW-k-means emphasized view 12; while our AMVFCM-U/AAMVFCM-U prioritized view 1. This differential view prioritization suggests these algorithms may be detecting complementary aspects of cellular identity, with each view potentially capturing distinct regulatory programs involved in retinal cell type specification. The multi-algorithm consensus around specific views (particularly 1, 7, 8, and 12) may identify core transcriptional modules defining the major axes of variation in retinal cell identity.

These findings have significant implications for transcriptomic analysis of complex tissues. The superior performance of kernel-based methods suggests that future scRNA-seq clustering approaches should explicitly model the non-linear manifold structure of cellular differentiation landscapes. Furthermore, the demonstrated feasibility of substantial dimensionality reduction (up to 54.5% within specific views) without compromising clustering accuracy suggests that many scRNA-seq datasets contain considerable redundancy that can be eliminated to improve both computational efficiency and biological interpretability.

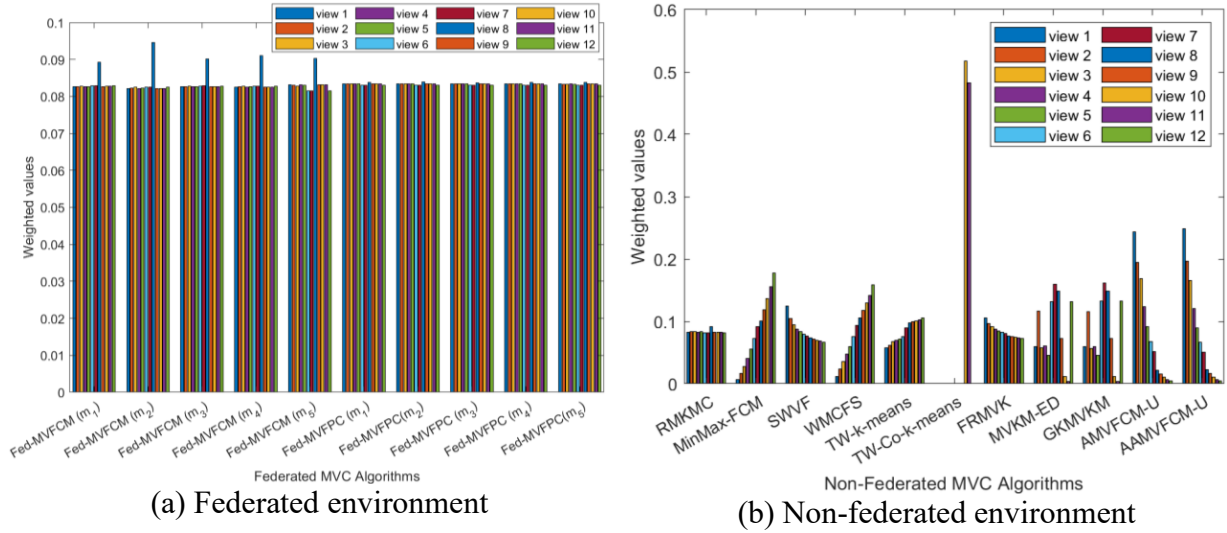

(a) Federated environment

(b) Non-federated environment

Figure 10: The view weights on Macosko data

Table 16: Feature dimensionality reduction achieved across views in the Macosko scRNA-seq dataset

| View | Method | | | | Reduction |
|------|--------|--------|--------|--------------|-----------|
| *(s)* | AMVFCM | AAMVFCM-U | FRMVK | Co-FW-MVFCM | % |
| *High Reduction Views (>40%)* | | | | | |
| 3 | 156 | 91 | 154 | - | **41.7%** |
| 8 | 156 | 71 | 154 | - | **54.5%** |
| *Moderate Reduction Views (10-40%)* | | | | | |
| 1 | 156 | 138 | 155 | - | 11.5% |
| 2 | 156 | 125 | 155 | - | 19.9% |
| 7 | 156 | 136 | 153 | - | 12.8% |
| Low Reduction Views (<10%) | | | | | |
| 4 | 156 | 147 | 155 | - | 5.8% |
| 5 | 156 | 150 | 155 | - | 3.8% |
| 6 | 156 | 144 | 156 | - | 7.7% |
| 9 | 156 | 150 | 155 | - | 3.8% |
| 10 | 156 | 150 | 155 | - | 3.8% |
| 11 | 156 | 149 | 155 | - | 4.5% |

| 12 | 156 | 144 | 156 | - | 7.7% |
|---|---|---|---|---|---|
| **Total** | **1.872** | **1.445** | **1.858** | **-** | **22.8%** |

Notes: AAMVFCM-U demonstrates superior dimensionality reduction across all views compared to both AMVFCM-U (no reduction) and FRMVK (minimal reduction). Views 3 and 8 show particularly dramatic feature reduction (41.7% and 54.5% respectively), suggesting these views contain substantial redundant or non-informative features for cell type classification. This significant dimensionality reduction contributes to AAMVFCM-U's computational efficiency while maintaining clustering performance on the 39-class Macosko retinal cell dataset.

Table 17:   Total runtime comparison across 17 multi-view clustering algorithms (in seconds)

| Algorithm | MV Data | | | | |
|---|---|---|---|---|---|
| | Synthetic | Prokaryotic phyla | MSRC-v1 | Human Kidney | Macosko |
| **Traditional MVC** | | | | | |
| Co-FKM | 6686.822 | 82.517 | 2600.415 | 83627.621 | 117250.745 |
| RMKMC | 23984.608 | 51.813 | 89.111 | 460.416 | 3194.945 |
| MultiNMF | 1003.326 | 48.365 | 2243.107 | 2523.895 | 8873.618 |
| **Feature/View Weighted** | | | | | |
| SWVF | 511.162 | 223.387 | 322.943 | 920.604 | 15701.569 |
| WMCFS | 515.745 | 254.740 | 320.362 | 908.343 | 21581.661 |
| TW-k-means | 540.133 | 252.300 | 316.287 | 919.986 | 23260.636 |
| TW-Co-k-means | 574.632 | 111.324 | 71.741 | 1187.673 | 5075.916 |
| WV-Co-FCM | 7529.833 | 1802.678 | 1675.574 | 46289.843 | 58604.904 |
| MinMax-FCM | 4399.878 | <u>4.021</u> | 200.069 | **31.570** | **110.962** |
| **Feature Reduction** | | | | | |
| FRMVK | <u>130.773</u> | *6.194* | **18.665** | 685.876 | 15185.479 |
| Co-FW-MVFCM | 152747.042 | 2118.544 | 2759.663 | 35549.198 | - |
| **Advanced/Federated** | | | | | |
| Fed-MVFCM | 528.191 | 87.429 | 145.600 | 1521.179 | 13714.887 |
| Fed-MVFPC | 354.253 | 39.386 | 67.702 | 536.010 | <u>963.265</u> |
| MVKM-ED | 2207.280 | 77.936 | 81.802 | 7197.348 | 41046.324 |
| GKMVKM | 1692.662 | 65.633 | 85.475 | 7222.17 | 33391.398 |

| Proposed Methods | | | | |
|---|---|---|---|---|
| AMVFCM-U | **111.370** | 28.581 | *60.294* | *96.409* | 2639.466 |
| AAMVFCM-U | *266.760* | **2.298** | <u>45.866</u> | <u>69.167</u> | *2176.321* |
| **Ratio (max/min)** | **2,289×** | **1,077×** | <u>**156×**</u> | <u>**2,650×**</u> | **1,057×** |

Notes: Runtime measured in seconds across 50 trials (100 for synthetic data). **Bold** = fastest; <u>underlined</u> = second fastest; *italic* = third fastest. "-" indicates experiment did not complete after 24 hours. "Avg. Speedup" represents average runtime performance relative to AMVFCM-U (higher is better). "Ratio" shows computational cost disparity between slowest and fastest algorithms per dataset.

## 5. Conclusions

This study introduces a unified parameter-free framework for multi-view clustering through two complementary algorithms—AMVFCM-U and its extension AAMVFCM-U—that fundamentally transform the analysis of heterogeneous multi-view data. Our work addresses a critical gap in existing approaches by eliminating reliance on fuzzification parameters while simultaneously enhancing dimensionality reduction capabilities through principled mathematical foundations.

Our principal contributions encompass three interrelated methodological innovations: First, we develop a novel mathematical framework that substitutes traditional fuzzification parameters with adaptive entropy regularization terms. This approach automatically calibrates the balance between cluster separation and membership uncertainty while preserving probabilistic interpretability. The dual-level entropy components simultaneously optimize both view-level and feature-level contributions without requiring manual parameter tuning that plagues conventional fuzzy clustering methods.

Second, our introduced signal-to-noise ratio (SNR) based regularization approach $(\delta_j^h = \frac{\bar{x}_j^h}{(\sigma_j^h)^2})$ provides a theoretically grounded mechanism for feature importance assessment with provable convergence properties. This SNR-based formulation naturally emphasizes features with high information content while suppressing noisy or redundant dimensions, creating a self-calibrating system that adapts to data characteristics across diverse domains.

Third, the AAMVFCM-U algorithm implements a progressive dimensionality reduction system that operates simultaneously at both feature and view levels through adaptive thresholding $(\theta^{h^{(t)}} = \frac{d_h^{(t)}}{n})$ . This hierarchical approach systematically identifies and eliminates irrelevant information during optimization, enabling automatic pruning of entire views when warranted by the data structure.

Comprehensive empirical evaluation across five diverse benchmarks (synthetic data, Prokaryotic phyla, MSRC-v1, Human Kidney, and Macosko) demonstrates the exceptional performance of our proposed methods against 15 state-of-the-art clustering algorithms from both centralized and decentralized paradigms. The quantitative results reveal consistent patterns of superiority across multiple domains: AAMVFCM-U achieved unprecedented dimensionality reduction in the Prokaryotic phyla dataset (98.3% overall, reducing to merely 0.45% of original textual features) while simultaneously improving clustering accuracy (RI: 0.8976±0.0108). For genomic single-cell data, our approach demonstrated remarkable stability in delineating complex cellular hierarchies across both Human Kidney (NMI: 0.4353) and Macosko retinal (AR: 0.5801) datasets.

The computational advantages of our approach are particularly evident in high-dimensional scenarios, where AAMVFCM-U reduces execution time by up to 97\% compared to collaborative approaches. This efficiency stems directly from the progressive dimensionality reduction, with computational complexity decreasing from $O(t \cdot n \cdot c \cdot \sum_{h=1}^{s} d_h)$ to $O(t \cdot n \cdot c \cdot \sum_{h=1}^{s} d_h^{'})$ where $d_{h'} \ll d_h$ in many real-world applications.

While our framework substantially reduces parameter dependence, optimization within theoretically established bounds for $\beta \in [\frac{d_h}{n}, \frac{3d_h}{n}]$ and $\eta \in [0.0015, 0.025 \cdot \frac{d_{\min}}{d_{\max}}]$ remains important for peak performance. These bounds provide practical implementation guidance while maintaining the framework's adaptive capabilities across diverse data types.

Future research directions include: (1) extending our framework to fully decentralized and federated learning environments where privacy constraints limit data sharing; (2) developing completely adaptive regularization techniques to further reduce parameter dependence through meta-learning approaches; (3) exploring semi-supervised variants that can leverage limited labeled data to guide clustering in partially annotated domains; and (4) investigating theoretical connections between entropy-regularized clustering and information-theoretic principles in heterogeneous data fusion.

In conclusion, our parameter-free entropy-regularized framework with hierarchical feature selection represents a significant advancement in multi-view clustering, offering a unified approach that simultaneously addresses the challenges of parameter sensitivity, computational efficiency, and dimensionality reduction while maintaining or improving clustering quality across diverse application domains.

## CRediT authorship contribution statement

**Kristina P. Sinaga:** Conceptualization, Methodology, Writing – review & editing, Writing – original draft, Visualization, Software, Formal analysis, Data curation, Validation. **Sara Colantonio:** Resources, Writing – review & editing, Supervision, Project administration, Funding acquisition. **Miin-Shen Yang:** Methodology, Writing – review & editing, Supervision, Validation.

## Declaration of Competing Interest

The authors declare that they have no known competing financial interests or personal relationships that could have appeared to influence the work reported in this paper.

## Data availability

Data will be made available on https://github.com/KristinaP09/AAMVFCM-U .

## References

[1]     J.C. Bezdek, R. Ehrlich, W. Full, FCM: The fuzzy c-means clustering algorithm, Comput Geosci 10 (1984) 191–203. https://doi.org/https://doi.org/10.1016/0098-3004(84)90020-7.

[2]     X. Wang, Y. Wang, L. Wang, Improving fuzzy c-means clustering based on feature-weight learning, Pattern Recognit Lett 25 (2004) 1123–1132. https://doi.org/https://doi.org/10.1016/j.patrec.2004.03.008.

[3]     L. Jing, M.K. Ng, J.Z. Huang, An Entropy Weighting k-Means Algorithm for Subspace Clustering of High-Dimensional Sparse Data, IEEE Trans Knowl Data Eng 19 (2007) 1026–1041. https://doi.org/10.1109/TKDE.2007.1048.

[4]     K. Polat, Classification of Parkinson's disease using feature weighting method on the basis of fuzzy C-means clustering, Int J Syst Sci 43 (2012) 597–609. https://doi.org/10.1080/00207721.2011.581395.

[5]     Z. Jiang, T. Li, W. Min, Z. Qi, Y. Rao, Fuzzy c-means clustering based on weights and gene expression programming, Pattern Recognit Lett 90 (2017) 1–7. https://doi.org/https://doi.org/10.1016/j.patrec.2017.02.015.

[6]     H.-J. Xing, M.-H. Ha, Further improvements in Feature-Weighted Fuzzy C-Means, Inf Sci (N Y) 267 (2014) 1–15. https://doi.org/https://doi.org/10.1016/j.ins.2014.01.033.

[7]     R.J. Hathaway, Y. Hu, Density-Weighted Fuzzy c-Means Clustering, IEEE Transactions on Fuzzy Systems 17 (2009) 243–252. https://doi.org/10.1109/TFUZZ.2008.2009458.

[8]     E. C. Simões, F. de A. T. de Carvalho, Gaussian kernel fuzzy c-means with width parameter computation and regularization, Pattern Recognit 143 (2023) 109749. https://doi.org/https://doi.org/10.1016/j.patcog.2023.109749.

[9]     Y. Tang, Z. Pan, W. Pedrycz, F. Ren, X. Song, Viewpoint-Based Kernel Fuzzy Clustering With Weight Information Granules, IEEE Trans Emerg Top Comput Intell 7 (2023) 342–356. https://doi.org/10.1109/TETCI.2022.3201620.

[10]    J. Zhou, L. Chen, C.L.P. Chen, Y. Zhang, H.-X. Li, Fuzzy clustering with the entropy of attribute weights, Neurocomputing 198 (2016) 125–134. https://doi.org/https://doi.org/10.1016/j.neucom.2015.09.127.

[11]    S. Bickel, T. Scheffer, Multi-View Clustering, in: Proceedings of the Fourth IEEE International Conference on Data Mining, IEEE Computer Society, USA, 2004: pp. 19–26.

[12]    G. Cleuziou, M. Exbrayat, L. Martin, J.H. Sublemontier, CoFKM: A centralized method for multiple-view clustering, in: Proceedings - IEEE International Conference on Data Mining, ICDM, 2009: pp. 752–757. https://doi.org/10.1109/ICDM.2009.138.

[13]    L. Sun, L. Li, W. Xu, W. Liu, J. Zhang, G. Shao, A Novel Classification Scheme for Breast Masses Based on Multi-View Information Fusion, in: 2010 4th International Conference on Bioinformatics and Biomedical Engineering, 2010: pp. 1–4. https://doi.org/10.1109/ICBBE.2010.5517742.

[14]    B. Zineddin, Z. Wang, Y. Shi, Y. Li, M. Du, X. Liu, A multi-view approach to cDNA micro-array analysis, Int J Comput Biol Drug Des 3 (2010) 91–111. https://doi.org/10.1504/IJCBDD.2010.035237.

[15] U. Fang, M. Li, J. Li, L. Gao, T. Jia, Y. Zhang, A Comprehensive Survey on Multi-View Clustering, IEEE Trans Knowl Data Eng 35 (2023) 12350–12368. https://doi.org/10.1109/TKDE.2023.3270311.

[16] Y. Jiang, F.L. Chung, S. Wang, Z. Deng, J. Wang, P. Qian, Collaborative Fuzzy Clustering From Multiple Weighted Views, IEEE Trans Cybern 45 (2015) 688–701. https://doi.org/10.1109/TCYB.2014.2334595.

[17] B. Jiang, F. Qiu, L. Wang, Multi-view clustering via simultaneous weighting on views and features, Applied Soft Computing Journal 47 (2016) 304–315. https://doi.org/10.1016/j.asoc.2016.06.010.

[18] Y. Wang, L. Chen, Multi-view fuzzy clustering with minimax optimization for effective clustering of data from multiple sources, Expert Syst Appl 72 (2017) 457–466. https://doi.org/10.1016/j.eswa.2016.10.006.

[19] M.S. Yang, K.P. Sinaga, Collaborative feature-weighted multi-view fuzzy c-means clustering, Pattern Recognit 119 (2021). https://doi.org/10.1016/j.patcog.2021.108064.

[20] X. Hu, J. Qin, Y. Shen, W. Pedrycz, X. Liu, J. Liu, An Efficient Federated Multiview Fuzzy C-Means Clustering Method, IEEE Transactions on Fuzzy Systems 32 (2024) 1886–1899. https://doi.org/10.1109/TFUZZ.2023.3335361.

[21] Y. Li, X. Hu, S. Yu, W. Ding, W. Pedrycz, Y.C. Kiat, Z. Liu, A Vertical Federated Multi-View Fuzzy Clustering Method for Incomplete Data, IEEE Transactions on Fuzzy Systems (2025) 1–15. https://doi.org/10.1109/TFUZZ.2025.3526978.

[22] K. Zhou, Y. Zhu, M. Guo, M. Jiang, MvWECM: Multi-view Weighted Evidential C-Means clustering, Pattern Recognit 159 (2025) 111108. https://doi.org/https://doi.org/10.1016/j.patcog.2024.111108.

[23] M.N. Ahmed, S.M. Yamany, A.A. Farag, T. Moriarty, Bias field estimation and adaptive segmentation of MRI data using a modified fuzzy C-means algorithm, in: Proceedings. 1999 IEEE Computer Society Conference on Computer Vision and Pattern Recognition (Cat. No PR00149), 1999: pp. 250-255 Vol. 1. https://doi.org/10.1109/CVPR.1999.786947.

[24] A. and M.A.M. Masulli F. and Schenone, Fuzzy Clustering Methods for the Segmentation of Multimodal Medical Images, in: P.J.G. and K.J. Szczepaniak Piotr S. and Lisboa (Ed.), Fuzzy Systems in Medicine, Physica-Verlag HD, Heidelberg, 2000: pp. 335–350. https://doi.org/10.1007/978-3-7908-1859-8_15.

[25]   M.N. Ahmed, S.M. Yamany, N. Mohamed, A.A. Farag, T. Moriarty, A modified fuzzy c-means algorithm for bias field estimation and segmentation of MRI data, IEEE Trans Med Imaging 21 (2002) 193–199. https://doi.org/10.1109/42.996338.

[26]   W. Cai, S. Chen, D. Zhang, Fast and robust fuzzy c-means clustering algorithms incorporating local information for image segmentation, Pattern Recognit 40 (2007) 825–838. https://doi.org/https://doi.org/10.1016/j.patcog.2006.07.011.

[27]   C. Wang, W. Pedrycz, M. Zhou, Z. Li, Sparse Regularization-Based Fuzzy C-Means Clustering Incorporating Morphological Grayscale Reconstruction and Wavelet Frames, IEEE Transactions on Fuzzy Systems 29 (2021) 1826–1840. https://doi.org/10.1109/TFUZZ.2020.2985930.

[28]   C. Wang, W. Pedrycz, Z. Li, M. Zhou, Residual-driven Fuzzy C-Means Clustering for Image Segmentation, IEEE/CAA Journal of Automatica Sinica 8 (2021) 876–889. https://doi.org/10.1109/JAS.2020.1003420.

[29]   P. Wahlberg, G. Lantz, Methods for robust clustering of epileptic EEG spikes, IEEE Trans Biomed Eng 47 (2000) 857–868. https://doi.org/10.1109/10.846679.

[30]   P. Masulli, F. Masulli, S. Rovetta, A. Lintas, A.E.P. Villa, Fuzzy Clustering for Exploratory Analysis of EEG Event-Related Potentials, IEEE Transactions on Fuzzy Systems 28 (2020) 28–38. https://doi.org/10.1109/TFUZZ.2019.2910499.

[31]   B. Minasny, A.B. McBratney, FuzME version 3.0, Australian Centre for Precision Agriculture, The University of Sydney, Australia (2002).

[32]   D.-C. WANG, G.-L. ZHANG, X.-Z. PAN, Y.-G. ZHAO, M.-S. ZHAO, G.-F. WANG, Mapping Soil Texture of a Plain Area Using Fuzzy-c-Means Clustering Method Based on Land Surface Diurnal Temperature Difference, Pedosphere 22 (2012) 394–403. https://doi.org/https://doi.org/10.1016/S1002-0160(12)60025-3.

[33]   C. Wu, X. Guo, A Novel Single Fuzzifier Interval Type-2 Fuzzy C-Means Clustering With Local Information for Land-Cover Segmentation, IEEE J Sel Top Appl Earth Obs Remote Sens 14 (2021) 5903–5917. https://doi.org/10.1109/JSTARS.2021.3085606.

[34]   J.-L. Chen, J.-H. Wang, A new robust clustering algorithm-density-weighted fuzzy c-means, in: IEEE SMC'99 Conference Proceedings. 1999 IEEE International Conference on Systems, Man, and Cybernetics (Cat. No.99CH37028), 1999: pp. 90–94 vol.3. https://doi.org/10.1109/ICSMC.1999.823160.

[35] W. Wang, C. Wang, X. Cui, A. Wang, Improving Fuzzy C-Means Clustering Based on Adaptive Weighting, in: 2008 Fifth International Conference on Fuzzy Systems and Knowledge Discovery, 2008: pp. 62–66. https://doi.org/10.1109/FSKD.2008.160.

[36] A.H. Hadjahmadi, M.M. Homayounpour, S.M. Ahadi, Robust weighted fuzzy c-means clustering, in: 2008 IEEE International Conference on Fuzzy Systems (IEEE World Congress on Computational Intelligence), 2008: pp. 305–311. https://doi.org/10.1109/FUZZY.2008.4630382.

[37] M. Hashemzadeh, A. Golzari Oskouei, N. Farajzadeh, New fuzzy C-means clustering method based on feature-weight and cluster-weight learning, Appl Soft Comput 78 (2019) 324–345. https://doi.org/https://doi.org/10.1016/j.asoc.2019.02.038.

[38] I. Khan, Z. Luo, J.Z. Huang, W. Shahzad, Variable Weighting in Fuzzy k-Means Clustering to Determine the Number of Clusters, IEEE Trans Knowl Data Eng 32 (2020) 1838–1853. https://doi.org/10.1109/TKDE.2019.2911582.

[39] Y. Li, M. Du, W. Zhang, X. Jiang, Y. Dong, Feature Weighting-Based Deep Fuzzy C-Means for Clustering Incomplete Time Series, IEEE Transactions on Fuzzy Systems 32 (2024) 6835–6847. https://doi.org/10.1109/TFUZZ.2024.3466175.

[40] M. Kaushal, Q.M. Danish Lohani, O. Castillo, Weighted Intuitionistic Fuzzy C-Means Clustering Algorithms, International Journal of Fuzzy Systems 26 (2024) 943–977. https://doi.org/10.1007/s40815-023-01644-5.

[41] K.P. Sinaga, Rectified Gaussian kernel multi-view k-means clustering, (2024). http://arxiv.org/abs/2405.05619.

[42] X. Cai, F. Nie, H. Huang, Multi-view K-means clustering on big data, in: Proceedings of the Twenty-Third International Joint Conference on Artificial Intelligence, AAAI Press, 2013: pp. 2598–2604.

[43] J. Liu, C. Wang, J. Gao, J. Han, Multi-View Clustering via Joint Nonnegative Matrix Factorization, n.d. https://epubs.siam.org/terms-privacy.

[44] M.-S. Yang, K.P. Sinaga, Federated Multi-View K-Means Clustering, IEEE Trans Pattern Anal Mach Intell 47 (2025) 2446–2459. https://doi.org/10.1109/TPAMI.2024.3520708.

[45] X. Chen, X. Xu, J.Z. Huang, Y. Ye, TW-(k)-means: Automated two-level variable weighting clustering algorithm for multiview data, IEEE Trans Knowl Data Eng 25 (2013) 932–944. https://doi.org/10.1109/TKDE.2011.262.

[46] Y.M. Xu, C.D. Wang, J.H. Lai, Weighted Multi-view Clustering with Feature Selection, Pattern Recognit 53 (2016) 25–35. https://doi.org/10.1016/j.patcog.2015.12.007.

[47] G.Y. Zhang, C.D. Wang, D. Huang, W.S. Zheng, Y.R. Zhou, TW-Co-k-means: Two-level weighted collaborative k-means for multi-view clustering, Knowl Based Syst 150 (2018) 127–138. https://doi.org/10.1016/j.knosys.2018.03.009.

[48] M.S. Yang, K.P. Sinaga, A feature-reduction multi-view k-means clustering algorithm, IEEE Access 7 (2019). https://doi.org/10.1109/ACCESS.2019.2934179.

[49] N. Gothania, S. Kumar, Multi-View Fuzzy Clustering with Weighted Attributes and Views, in: 2018 Second International Conference on Intelligent Computing and Control Systems (ICICCS), 2018: pp. 353–359. https://doi.org/10.1109/ICCONS.2018.8663236.

[50] Y. Liu, J. Chen, Y. Lu, W. Ou, Z. Long, C. Zhu, Adaptively Topological Tensor Network for Multi-view Subspace Clustering, IEEE Trans Knowl Data Eng (2024). https://doi.org/10.1109/TKDE.2024.3391627.

[51] H. Yu, L. Jiang, J. Fan, S. Xie, R. Lan, A feature-weighted suppressed possibilistic fuzzy c-means clustering algorithm and its application on color image segmentation, Expert Syst Appl 241 (2024) 122270. https://doi.org/https://doi.org/10.1016/j.eswa.2023.122270.

[52] L. Hua, Y. Gu, X. Gu, J. Xue, T. Ni, A Novel Brain MRI Image Segmentation Method Using an Improved Multi-View Fuzzy c-Means Clustering Algorithm, Front Neurosci 15 (2021). https://doi.org/10.3389/fnins.2021.662674.

[53] Q. Rao, Y. Yang, Y. Jiang, Condition recognition of high-speed train bogie based on multi-view kernel FCM, Big Data Mining and Analytics 2 (2019) 1–11. https://doi.org/10.26599/BDMA.2018.9020027.

[54] W. Yiping, S. Buqing, W. Jianjun, W. Qing, L. Haowu, L. Zhanxiang, Z. Ningning, C. Qingchao, An improved multi-view collaborative fuzzy C-means clustering algorithm and its application in overseas oil and gas exploration, J Pet Sci Eng 197 (2021) 108093. https://doi.org/https://doi.org/10.1016/j.petrol.2020.108093.

[55] M. Gönen, E. Alpaydın, Multiple Kernel Learning Algorithms, 2011.

[56] A. Golzari Oskouei, N. Samadi, S. Khezri, A. Najafi Moghaddam, H. Babaei, K. Hamini, S. Fath Nojavan, A. Bouyer, B. Arasteh, Feature-weighted fuzzy clustering methods: An experimental review, Neurocomputing 619 (2025) 129176. https://doi.org/https://doi.org/10.1016/j.neucom.2024.129176.

[57] M.D.J. C., Information theory, inference and learning algorithms, Http://Wol.Ra.Phy.Cam.Ac.Uk/Mackay/ (n.d.). https://cir.nii.ac.jp/crid/1571135650672053504.bib?lang=en (accessed June 4, 2025).

[58] S. DASGUPTA, Learning mixture of Gaussians, Proc. 40th IEEE Symposium on Foundation on Computer Science, 1999 (1999). https://cir.nii.ac.jp/crid/1573387451535587200.bib?lang=en (accessed June 4, 2025).

[59] M. Brbić, M. Piškorec, V. Vidulin, A. Kriško, T. Šmuc, F. Supek, The landscape of microbial phenotypic traits and associated genes, Nucleic Acids Res 44 (2016) 10074–10090. https://doi.org/10.1093/nar/gkw964.

[60] J. Winn, A. Criminisi, T. Minka, Object Categorization by Learned Universal Visual Dictionary, 2005. http://research.microsoft.com/vision/cambridge/recognition/.

[61] M. Brbić, I. Kopriva, Multi-view low-rank sparse subspace clustering, Pattern Recognit 73 (2018) 247–258. https://doi.org/https://doi.org/10.1016/j.patcog.2017.08.024.

[62] J. Shotton, A. Blake, R. Cipolla, Contour-based learning for object detection, in: Tenth IEEE International Conference on Computer Vision (ICCV'05) Volume 1, 2005: pp. 503-510 Vol. 1. https://doi.org/10.1109/ICCV.2005.63.

[63] E.Z. Macosko, A. Basu, R. Satija, J. Nemesh, K. Shekhar, M. Goldman, I. Tirosh, A.R. Bialas, N. Kamitaki, E.M. Martersteck, J.J. Trombetta, D.A. Weitz, J.R. Sanes, A.K. Shalek, A. Regev, S.A. McCarroll, Highly parallel genome-wide expression profiling of individual cells using nanoliter droplets, Cell 161 (2015) 1202–1214. https://doi.org/10.1016/j.cell.2015.05.002.

[64] G. Yang, J. Zou, Y. Chen, L. Du, P. Zhou, Heat Kernel Diffusion for Enhanced Late Fusion Multi-View Clustering, IEEE Signal Process Lett 31 (2024) 2310–2314. https://doi.org/10.1109/LSP.2024.3449229.

[65] W.M. Rand, Objective Criteria for the Evaluation of Clustering Methods, 1971.

[66] P. JACCARD, Distribution de la flore alpine dans le Bassin des Dranses et dans quelques regions voisines, Bull Soc Vaudoise Sci Nat 37 (1901) 241–272. https://cir.nii.ac.jp/crid/1570009751307275008.bib?lang=en (accessed February 19, 2025).

[67] E.B. Fowlkes, C.L. Mallows, A Method for Comparing Two Hierarchical Clusterings, 1983.

[68]   T.M. Cover, J.A. Thomas, Elements of Information Theory (Wiley Series in Telecommunications and Signal Processing) (Hardcover), n.d.

[69]   D.H. Parks, M. Chuvochina, D.W. Waite, C. Rinke, A. Skarshewski, P.-A. Chaumeil, P. Hugenholtz, A standardized bacterial taxonomy based on genome phylogeny substantially revises the tree of life, Nat Biotechnol 36 (2018) 996–1004. https://doi.org/10.1038/nbt.4229.

[70]   Q. Zhu, U. Mai, W. Pfeiffer, S. Janssen, F. Asnicar, J.G. Sanders, P. Belda-Ferre, G.A. Al-Ghalith, E. Kopylova, D. McDonald, T. Kosciolek, J.B. Yin, S. Huang, N. Salam, J.-Y. Jiao, Z. Wu, Z.Z. Xu, K. Cantrell, Y. Yang, E. Sayyari, M. Rabiee, J.T. Morton, S. Podell, D. Knights, W.-J. Li, C. Huttenhower, N. Segata, L. Smarr, S. Mirarab, R. Knight, Phylogenomics of 10,575 genomes reveals evolutionary proximity between domains Bacteria and Archaea, Nat Commun 10 (2019) 5477. https://doi.org/10.1038/s41467-019-13443-4.

[71]   X. Gao, H. Liang, T. Hu, Y. Zou, L. Xiao, Cultivated genome references for protein database construction and high-resolution taxonomic annotation in metaproteomics, Microbiol Spectr 13 (2025) e01755-24. https://doi.org/10.1128/spectrum.01755-24.

[72]   K. Suhre, M.I. McCarthy, J.M. Schwenk, Genetics meets proteomics: perspectives for large population-based studies, Nat Rev Genet 22 (2021) 19–37. https://doi.org/10.1038/s41576-020-0268-2.